\documentclass[acmsmall, screen, nonacm, authorversion]{acmart}
\usepackage{colortbl}
\usepackage{multirow}
\usepackage{subfig}
\usepackage[ruled]{algorithm2e}
\usepackage{amsmath}
\usepackage{graphics}
\usepackage{soul}

\AtBeginDocument{%
  \providecommand\BibTeX{{%
    \normalfont B\kern-0.5em{\scshape i\kern-0.25em b}\kern-0.8em\TeX}}}

\begin{document}

\title{Sense and Learn: Self-Supervision for Omnipresent Sensors}

\author{Aaqib Saeed}
\email{a.saeed@tue.nl}
\affiliation{
  \institution{Eindhoven University of Technology}
  \city{Eindhoven}
  \country{The Netherlands}
}
\authornote{This work was conducted at Google.}
\author{Victor Ungureanu}
\email{ungureanu@google.com}
\affiliation{
    \institution{Google Research}
    \city{Zurich}
    \country{Switzerland}
}
\author{Beat Gfeller}
\email{beatg@google.com}
\affiliation{
    \institution{Google Research}
    \city{Zurich}
    \country{Switzerland}
}

\begin{abstract}
Learning general-purpose representations from multisensor data produced by the omnipresent sensing systems (or IoT in general) has numerous applications in diverse use cases. Existing purely supervised end-to-end deep learning techniques depend on the availability of a massive amount of well-curated data, acquiring which is notoriously difficult but required to achieve a sufficient level of generalization on a task of interest. In this work, we leverage the self-supervised learning paradigm towards realizing the vision of continual learning from unlabeled inputs. We present a generalized framework named \textit{Sense and Learn} for representation or feature learning from raw sensory data. It consists of several auxiliary tasks that can learn high-level and broadly useful features entirely from unannotated data without any human involvement in the tedious labeling process. We demonstrate the efficacy of our approach on several publicly available datasets from different domains and in various settings, including linear separability, semi-supervised or few shot learning, and transfer learning. Our methodology achieves results that are competitive with the supervised approaches and close the gap through fine-tuning a network while learning the downstream tasks in most cases. In particular, we show that the self-supervised network can be utilized as initialization to significantly boost the performance in a low-data regime with as few as $5$ labeled instances per class, which is of high practical importance to real-world problems. Likewise, the learned representations with self-supervision are found to be highly transferable between related datasets, even when few labeled instances are available from the target domains. The self-learning nature of our methodology opens up exciting possibilities for on-device continual learning.
\end{abstract}


\maketitle

\section{Introduction}
The rise of deep neural networks for learning general-purpose representations in an end-to-end manner has led to numerous breakthroughs in different areas of artificial intelligence, including object recognition~\cite{ren2015faster}, complex gameplay~\cite{silver2017mastering}, and language modeling~\cite{devlin2018bert}. These advancements have brought their widespread adoption to other domains, particularly for problems involving time-series or sensory inputs, which, crucially, depended on ad-hoc feature extraction with shallow learning techniques. The efficiency of deep learning algorithms substantially improved the state-of-the-art in these fields~\cite{supratak2017deepsleepnet, martinez2013learning, hannun2019cardiologist, saeed2018model, radu2018multimodal}; while largely dismissing manual feature design strategies. However, this success is due to supervised learning models, which require a huge amount of well-curated data to solve the desired task.
Compared to computer vision or other realms, semantically-labeled sensory data (such as electrooculography, heart rate variability, and inertial signals) is much more difficult to acquire, owing to: privacy issues, complicated experimental set-ups and the prerequisite of expert-level knowledge for data labeling.

Due to these limitations, unsupervised learning holds an enormous potential to leverage a vast amount of unlabeled data produced via omnipresent sensing systems. For instance, an average smartphone or smartwatch is equipped with a multitude of sensors, such as IMUs, microphone, proximity, ambient light and heart rate monitors producing a wealth of data that can be utilized for solving challenging problems and can enable novel use cases through harnessing the power of machine learning. Past efforts to learn from sensory (or time-series) data were mainly limited to the use of autoencoding based approaches~\cite{li2014unsupervised, bhattacharya2014using, martinez2013learning, plotz2011feature} that can learn to compress the data, but fail to learn semantically useful features~\cite{oord2018representation}. More recently, generative adversarial networks (GANs) have been explored to some extent for unsupervised learning from sensory inputs~\cite{yao2018sensegan}, but GANs are infamous for being notoriously unstable during training and suffer from mode collapse, making it a great challenge to use them in practice, for now~\cite{thanh2019improving}. It might also be excessive to use GANs as a pre-training strategy when synthesizing data is not a core focus, as the number of parameters in the network that need to be learned increases extensively. Moreover, transfer learning has been utilized to a limited extent for tackling the issue of unavailability of massive well-annotated sensory datasets for training deep models. It has been explored to improve the performance in a supervised setting through joint-training on labeled source and target datasets~\cite{chen2019cross, gjoreski2019cross}. In these cases, the features transferred from supervised models may not be general and are mostly tied to a specific task; therefore, they might not generalize well to other tasks of interest, compared to methods that learn task-agnostic features, in an unsupervised manner. Likewise, existing methods did not focus on learning in low-data regimes nor from unlabeled input which is available in much larger quantities (see section~\ref{sec:rw} for related work). In this paper, we show that the  emerging paradigm of self-supervised learning offers an efficient way for learning semantically-meaningful representations from sensory data that can be used for solving a diverse set of downstream tasks~\footnote{downstream or end tasks referred to the tasks of interest e.g., sleep stage scoring.}. The self-supervised approaches exploit the inherent structure of the input to derive a supervisory signal. The idea is to define a pretext task, for which annotations can be acquired without human involvement (directly from the raw data) and can be solved using some form of unsupervised learning techniques. This intriguing property essentially renders a deep sensing model, that is developed based on the earlier described principle of "self-learning" in nature: a system that can be trained continuously on massive, readily-accessible data in an unsupervised manner~\cite{de1994learning, schmidhuber1990making}. However, in this case, the challenge lies in designing complex auxiliary tasks that can force the deep neural network to capture meaningful features of the input, while avoiding shortcuts~\cite{geirhos2020shortcut} (i.e., simple unintended ways to trivially solve the auxiliary task without learning anything useful that generalizes beyond the auxiliary task). 

Over the last few years, given the large potential of self-supervised learning in exploiting unlabeled data, multiple surrogate or auxiliary tasks have been proposed for feature learning to ultimately solve complex problems in different domains~\cite{oord2018representation, gidaris2018unsupervised, devlin2018bert}. Particularly in the vision community, a surge has been seen in developing self-supervised methods, owing to the availability of a wide variety of large scale datasets and well-established deep network architectures. In this realm, the most straightforward strategy is the reconstruction of contextual information based on partially observable input~\cite{doersch2015unsupervised}. The prediction of color values for grayscale images~\cite{zhang2016colorful} and the detection of the angle of rotation~\cite{gidaris2018unsupervised} are recent attempts found to be useful in learning visual representations. Similarly, the temporal synchronization of multimodal data is exploited to learn audio-visual representations~\cite{korbar2018cooperative}. Likewise, contrastive learning is another highly promising technique that aim to capture shared information among multiple views of the data~\cite{tian2019contrastive, oord2018representation}, including successes in robotic imitation learning~\cite{Sermanet2017TCN}. Thus, we conjecture that self-supervision is fruitful for automatically extracting generic latent embeddings from sensory data that can improve much-needed label efficiency, as acquiring well-labeled sensory data is extremely challenging in the real world. Furthermore, due to its annotation-free nature, this learning strategy is not only effective and scalable, but can also be directly leveraged in a federated learning environment~\cite{bonawitz2019towards}, to learn from widely distributed and decentralized data without aggregating it in a centralized repository, which can preserve users' privacy~\cite{mcmahan2017communication}. 

In this paper, we present a principled framework for self-supervised learning of multisensor representations from unlabeled data. Our objective is to have numerous tasks, with each perhaps imposing a distinct prior on to the learning process, resulting in varying quality features that may differ across sensing datasets. Specifically, as proxy tasks and modalities could be of more or less relevance to the downstream task's performance, it is essential to explore and compare several pretext tasks so as to discover the ones with better generalization properties. The broad aim is to have many auxiliary tasks in a user's toolbox such that, either experimentally or based on prior knowledge, a relevant task can be selected for training deep models. Particularly, the objective is to have proxy tasks that enable learning of representations invariant to several input deformations that commonly arise in the temporal data, such as sensor noise and sampling-rate disparities, or that can be used jointly in a multi-task learning setting. To this end, we develop eight novel auxiliary tasks that intrinsically obtain supervision from the unlabeled input signals to learn general-purpose features with a temporal convolutional network, such that the pre-trained model generalizes well to the end tasks.

\begin{figure}[t]
\centering
\includegraphics[width=10.5cm]{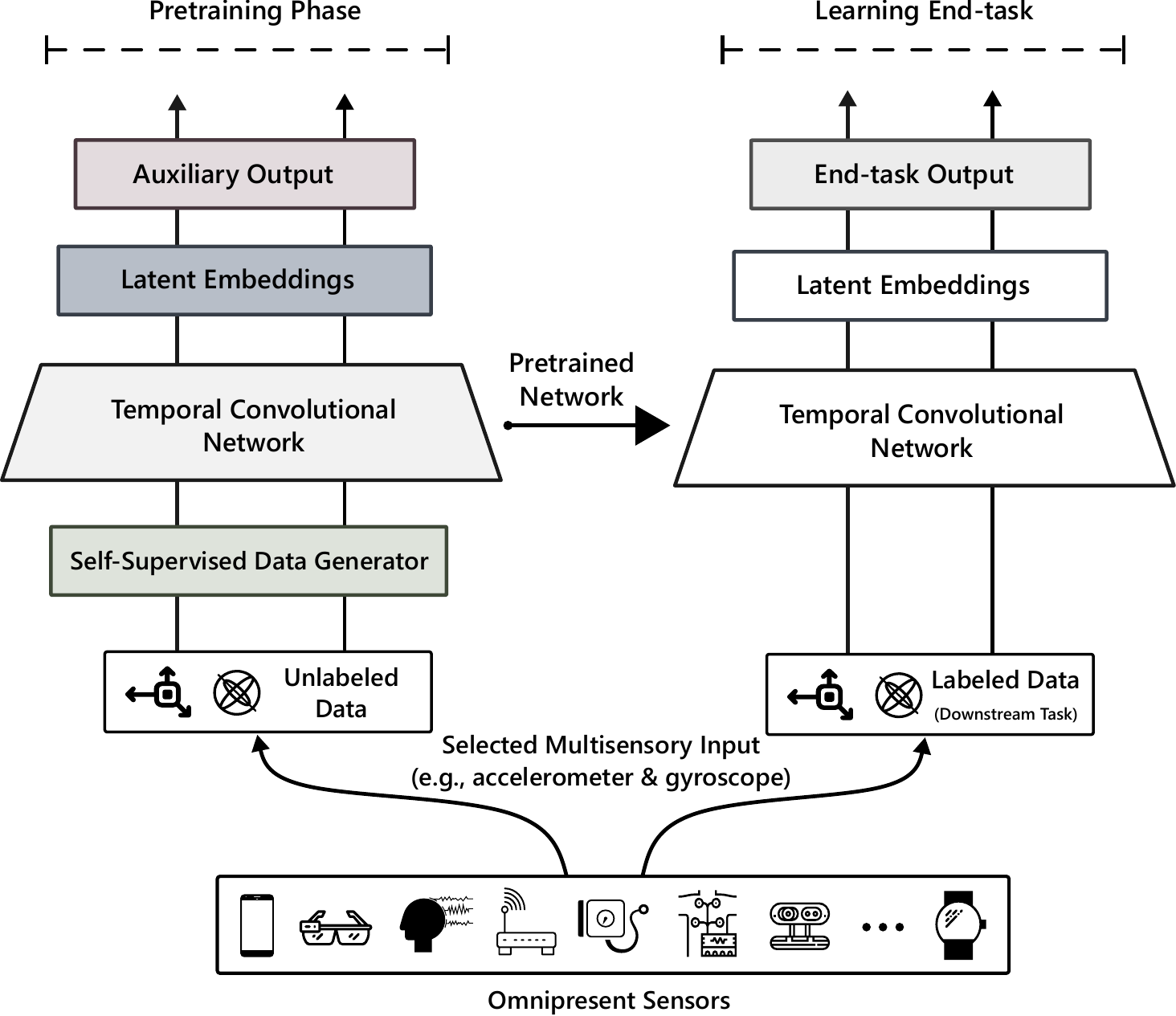}
\caption{Illustration of our \textit{Sense and Learn} representation learning framework. A deep neural network is pre-trained with self-supervision using input modalities from large unlabeled sensory data, such as inertial measurements (or electroencephalogram, heart rate, and channel state information). The learned network can then be utilized as a feature-extractor or initialization for rapidly solving downstream tasks of interest with few labeled data.}
\label{fig:overview}
\end{figure}

Our approach comprises of pre-training a network through self-supervision with unlabeled data so that it captures high-level semantics and can be used either as a feature extractor\footnote{i.e. leveraging representations from intermediate layers of the deep neural network} or utilized as initialization for making successive tasks of interest easier to solve with few labeled data. To develop the auxiliary tasks, we take advantage of the synchronized multisensor (or multimodal) data as it belongs to the same underlying phenomena and we exploit it to create proxy tasks that can capture broadly useful features. Specifically, it can substantially help in learning powerful representations of each modality, and ultimately learn more abstract concepts in a joint-embedding space. Thus, we use a multi-stream neural network architecture to solve proxy tasks so that it can learn modality-specific features with a distinct encoder per modality and subsequently learn a shared embedding space with a modality-agnostic encoder. The fundamental structure of our framework is illustrated in Figure~\ref{fig:overview}. We adopt a small model architecture in this work to highlight a) effectiveness of self-supervised tasks (i.e. improvement is not due to complex architecture) and b) potential of deployment on resource-constrained devices for training and inference.

We demonstrate that a relatively straightforward suite of auxiliary tasks results in meaningful features for diverse problems, including: activity recognition, stress detection, sleep stage scoring, and WiFi sensing. First, we show that the self-supervised representations are highly competitive with those learned with a fully-supervised model, by training a linear classifier on top of the frozen network, as it is a standard evaluation protocol for assessing the quality of self-supervised tasks~\cite{tagliasacchi2019self, oord2018representation}. Second, we explore fine-tuning the last layer of the encoder to gain further improvements over training from scratch. Third, we investigate the effectiveness of the learned representations in low-data regime\footnote{or in a semi-supervised setting}. Using our pre-trained network as initialization, we achieve a significant performance boost with as little as $5$ to $10$ labeled instances per class, which clearly highlights the value of self-supervised learning. Lastly, we evaluate the transferability of the features across related datasets/tasks to show the generality of our method in an unsupervised transfer learning setting.

In summary, our main contributions are as follows:

\begin{itemize}
\item We propose \textit{Sense and Learn}, a generalized self-supervised learning framework comprising several surrogate tasks to extract semantic structural concepts inherent to diverse types of sensory or time-series data. 

\item We extensively evaluate our self-supervised tasks on various problems (e.g. sleep stage scoring, activity recognition, stress detection, and WiFi sensing) and learning settings (i.e. transfer and semi-supervised) to significantly improve the data efficiency or lower the requirement of collecting large-scale labeled datasets.

\item Our results demonstrate that self-supervision provides an effective initialization of the network (and powerful embeddings) that improves performance significantly with minimal fine-tuning, and works well in a low-data regime, which is of high importance for real-world use cases.

\item The developed auxiliary tasks require an equivalent computational cost as standard supervised learning and has fewer parameters than autoencoding methods, but provide better generalization with greatly improved sample efficiency. 

\item We utilize a small network architecture to show the capability of self-supervision and its prospective usage on resource-constrained devices. In particular, the majority of our proposed tasks are designed around the principle that self-supervised data generation should not be computationally expensive; thus, it can be readily used for on-device learning. 

\item We briefly discuss how to use our framework in practice, as well as its limitations.

\end{itemize}

\noindent In the following sections, we present the relevant literature to our work in Section~\ref{sec:rw}. Our self-supervised methodology is described in Section~\ref{methodology}. The experimental results are discussed in Section~\ref{experiments}, real-world impact and limitations in Section~\ref{sec:impact}, and conclusions and directions for future work are presented in Section~\ref{sec:conclusion}.

\section{Related Work}
\label{sec:rw}
\subsection{Unsupervised and Self-Supervised Learning}
Deep learning has revolutionized several areas of research with an intuitive property of learning discriminative features directly from the raw data and eliminating the need of manual feature extraction\cite{radu2018multimodal, hammerla2016deep, martinez2013learning, hannun2019cardiologist}. The success of deep learning is largely attributed to the massive labeled datasets apart from other factors, such as availability of computational power and better neural architectures. Obtaining semantically labeled data required for training supervised models is an expensive and time-consuming process. Therefore, unsupervised learning has seen growing interest in the last couple of years as unlabeled data is available in huge quantities, especially on decentralized edge devices.  A classical illustration of unsupervised feature learning is the autoencoder, which learns to map an input onto a lower-dimensional embedding so that reconstructing the original input from such a space incurs a lower error. However, the decoding-based strategies deplete the network capacity through attending to low-level details instead of capturing semantically meaningful features. Therefore, the focus of recent studies is on providing an alternative form of supervision, where annotations can be intrinsically extracted from the data itself. 

The field of self-supervised learning exploits the natural supervision available within the input signal to define a surrogate task that can force the network to learn broadly-usable representations. To that end, numerous pretext tasks are proposed in different domains.~\cite{noroozi2016unsupervised} established the task of predicting the relative position of randomly cropped image patches.~\cite{larsson2016learning, zhang2016colorful} inferred color values for grayscale pictures,~\cite{Sermanet2017TCN} utilize time-contrastive loss as a way to minimize the embedding distances of the same scene recorded from multiple viewpoints, while maximizing the distances for those captured at different timesteps. A similar technique is proposed in~\cite{tian2019contrastive} to learn from multiple views of the data.~\cite{tagliasacchi2019self} defined self-supervised tasks for audio, inspired by word$2$vec~\cite{mikolov2013distributed}.~\cite{korbar2018cooperative} showed that video representations could be learned by exploiting audio-visual temporal synchronization. Time-contrastive learning is suggested in~\cite{hyvarinen2016unsupervised} for extracting features from time-series, in an unsupervised manner, through predicting segment IDs. Likewise, autoregressive modeling has been combined with predictive coding to learn compact latent embeddings for various domains~\cite{oord2018representation}. For natural language modeling, self-supervised objectives, such as predicting masked tokens from surrounding ones and predicting the next sentence, turn out to be powerful methods for learning generic representations of text~\cite{devlin2018bert}. Similarly, for learning inertial sensory features, ~\cite{saeed2019multi} presented a signal transformation recognition task. Lately, self-supervised learning has been shown to be beneficial for semi-supervised learning, through jointly optimizing supervised and self-supervised losses~\cite{zhai2019s}. In this work, we develop several self-supervised tasks for learning representations from a wide range of sensory data such as electroencephalography, electrodermal activity and inertial signals. We show that pre-training with self-supervision using unlabeled data helps in learning highly generalizable features that improve data efficiency and transfer well to a related set of tasks.  

\subsection{Learning Sensing Models with Machine Learning}
An understanding of human contexts, activities and states is an important area of research in ambient computing and pervasive sensing due to the fact that it can play a central role in several application domains including: health, wellness, assistance, monitoring, and human computer interaction. To achieve the earlier described objective, the data is collected from users through wearables or other sensors, under varied environments, for learning a task-specific model. For instance, prior work on activity recognition explored various methodologies with inertial sensors embedded in smartphones or smartwatches~\cite{himberg2001time, stisen2015smart, hammerla2016deep}. Emotional state recognition is widely achieved with physiological signals, such as skin conductance and heart rate variability~\cite{saeed2018model, martinez2013learning, picard2001toward}. Similarly in sleep analysis, the electrical brain activity is captured with an electroencephalogram to classify sleep into different stages~\cite{supratak2017deepsleepnet, lajnef2015learning, gunecs2010efficient}. Importantly, for device-free sensing systems, channel state information from WiFi is utilized to infer participants' activities in a non-intrusive manner~\cite{yousefi2017survey}. Earlier developed methods for these problems heavily relied on manual feature extraction from sensory data to infer a user's activity, emotional state or sleep score and these methods were limited depending on the domain knowledge available to extract discriminative features.  With the tremendous progress in end-to-end supervised learning via deep networks, it has been shown that the features can be learned directly from data instead of hand-crafting them based on domain knowledge \cite{radu2018multimodal, hammerla2016deep, martinez2013learning, hannun2019cardiologist}.

Consequently, 1D convolutional and recurrent neural networks have become standard techniques for achieving state-of-the-art performance on problems involving temporal data~\cite{hannun2019cardiologist, saeed2018model, supratak2017deepsleepnet, hammerla2016deep}. Nevertheless, these approaches have heavily relied on the availability of large-annotated datasets, which are notoriously difficult to acquire in the real-world. Due to this, in recent years, few work explored unsupervised feature learning to exploit the availability of vast amounts of unlabeled data, while mainly focusing on input reconstruction via autoencoders and related variants, such as restricted Boltzmann machines and sparse coding ~\cite{li2014unsupervised, bhattacharya2014using, martinez2013learning, plotz2011feature}. There has also been work on utilizing generative adversarial networks for modeling data distributions without supervision~\cite{luo2018multivariate, esteban2017real} and in semi-supervised learning for sensing models~\cite{yao2018sensegan}. Furthermore, transfer learning has also been leveraged to improve neural network generalization in domains where large labeled data is difficult to obtain, but focused on transfer from supervised models~\cite{chen2019cross, gjoreski2019cross}. More recently,~\cite{saeed2019multi} proposed a self-supervised task of signal transformation recognition for feature learning that achieved significant improvement in activity recognition over autoencoding, though focusing only on unimodal input and the activity recognition problem. As opposed to earlier works, we present a general framework for learning multimodal representations from a diverse set of sensors in a self-supervised way and compared to~\cite{saeed2019multi} we simplify the problem formulation of transformation recognition (see section~\ref{sec:sslt}); our novel proxy tasks work on-par and can be used when transforming the input is not desirable or when it may lead to unintended outcomes (e.g. ECG signals). Furthermore, pre-training models with our auxiliary tasks significantly lower the amount of labeled data required to achieve better generalization and opens up the possibility of on-device learning from decentralized unlabeled data.

\section{Methodology}
\label{methodology}
In this section, we begin with a motivation and an overview of our self-supervised framework for learning sensory representations. Next, we provide a formalization of the auxiliary tasks and discuss an end-to-end approach for mutli-modal learning. Subsequently, we describe the network architecture design, its implementation, and the optimization procedure. 

\subsection{Motivation and Overview}
The key insight behind our technique is that the self-supervised pre-training acts as a prior that can give rise to varying quality representations that encode underlying signal semantics at different levels, which may or may not be useful for a downstream-task of interest. Therefore, it is vital to employ multiple auxiliary tasks to discover the suitable inductive bias necessary to obtain optimal performance on the desired end-task. This intuition is important considering that the time-series (or sensory) data shows peculiar characteristics (e.g. signal-to-noise ratio, amplitude variances, and sampling rates) depending on the nature of phenomena being recorded. Likewise, there should be an array of tasks to choose from depending on the learning problem and device type (e.g. available resources, sensor types etc.). Importantly, we want the self-supervised model to learn generic features rather than focusing on low-level input details, as a pre-trained network has to provide a strong initialization for learning with limited labeled data and generalize to other related tasks. Thus, instead of relying on a single auxiliary task, we learn latent representations with a broad set of tasks based on different objective functions.   

We propose a generalized framework comprising of eight pretext tasks that can be used to learn features from heterogeneous multisensor data. To achieve this, we utilize a temporal convolutional network (TCN) $F_\theta$ with a distinct encoder $e_m$ for each input modality $I_m$ and a shared encoder $e_s$ for multi-modal representation extraction. We choose to use TCN as an embedding network for sequence modeling due to its effectiveness in capturing long-term dependencies and parallelizability at a significantly lower cost than recurrent networks~\cite{bai2018empirical}. For every learning problem, we consider unlabeled multisensor (or multimodal) data $\mathcal{D} = \{(\textbf{u}_1, \textbf{v}_1), (\textbf{u}_2, \textbf{v}_2), \ldots (\textbf{u}_n, \textbf{v}_n)\}$ consisting of $N$ examples. Here, $\textbf{u}_n$ and $\textbf{v}_n$ denote the samples of different modalities (e.g. accelerometer and gyroscope) of the $n^{th}$ example. The defined pretext tasks exploit the inherent properties of the data to obtain supervision from the input pairs without requiring any manual annotation to optimize a certain loss function. Specifically, each surrogate task employs its own loss function $L_t$ for learning $F_\theta$ differently. For instance, an input reconstruction task employs mean-square error loss, while another task, concerning the detection of odd segments within a signal, uses negative log-likelihood; we discuss these in detail in the subsequent section. At a high-level, we utilize these objectives as necessary proxies for sensory representation learning without focusing on how well the model performs on them but on an end-task. After pre-training, $F_\theta$ captures a joint embedding space of the inputs, and thus it can be utilized either as a feature extractor or as initialization for rapidly learning to solve other problems. Finally, it is important to note that proxy tasks cannot be applied arbitrarily to any type of input and tasks like blend detection can only be used when modalities are related to each other, e.g. as accelerometer and gyroscope.

\subsection{Self-Supervised Tasks}
\label{sec:sslt}
In order to achieve self-supervised learning of disentangled semantic representations from unannotated sensory data, we develop eight surrogate tasks for the network. To solve these tasks, we assume $\textbf{u} = \{u_1, u_2, \ldots, u_l\}$ and $\textbf{v} = \{v_1, v_2, \ldots, v_l\}$ denote multi-channel signals of length $l$ from different modalities (e.g. accelerometer and gyroscope). Let $z_u = e_u(\textbf{u})$ and $z_v = e_v(\textbf{v})$ be the low-dimensional embeddings computed from the corresponding input signals with respective encoders. Likewise, $z_s = e_s(e_u(\textbf{u}), e_v(\textbf{v}))$ provides a shared embedding of the inputs through fusion that may capture more abstract features. A high-level illustration of the self-supervised learning procedure is shown in Figure~\ref{fig:overview}. A self-supervised data generation module produces annotated input from unlabeled multisensor data for learning $F_\theta$. We utilize this formulation to define the self-supervised objectives in the following subsections.

\subsection*{Blend Detection}
To take advantage of the multisensor signals, we define an auxiliary task of detecting input blending as a multi-class classification problem. Given an unlabeled input batch $B = \cup_{i=1}^{|B|} \{(\textbf{u}, \textbf{v})\}_i$, we generate three types of instances. First, we keep the original samples as belonging to a class $c_a$. Second, we perform a weighted blending of an instance from one modality with another randomly selected example from a different modality as class $c_b$. Third and last, the instances of the same modalities are blended to have instances for a class $c_c$. The blending weight $\mu$ is sampled from a uniform distribution, i.e. $\mu \sim \mathcal{U}(0, 1)$. The network is trained with a negative log-likelihood loss $\mathcal{L}_{NL}$ for learning to differentiate between examples of blended and clean classes ($y_k$) on the entire training set $\mathcal{D}_{train}$:

\begin{align*} 
\mathcal{L}_{NL} =  - \frac{1}{K} \sum_{k=1}^{K} y_{k} \times \log(F_\theta(\textbf{u},\textbf{v}))
\end{align*}

\subsection*{Fusion Magnitude Prediction}
We create a variant of the earlier defined task that uses a similar data generation strategy but differs fundamentally in terms of the objective it optimizes. The data generation process is similar to the mixup strategy, widely used for data augmentation in supervised learning~\cite{zhang2017mixup}. Here, we task the network with predicting the magnitude $\mu$, which defines the blending (or weighting) factor of the signals. We assign $\mu = 0$ to the clean examples, while assigning weight $\mu \sim \mathcal{U}(0, 1)$ to the blended examples, as earlier. In this case, a natural choice is to adopt mean-square loss as learning objective. However, we experimentally discovered that utilizing binary cross-entropy with a logistic function in the network's output layer results in better generalization; thus the network is trained to minimize the following loss $\mathcal{L}_{BCE}$ for each input modality:   

\begin{align*} 
\mathcal{L}_{BCE} = -(y \times \log(F_\theta(\textbf{u}, \textbf{v})) + (1 - y) \times \log(1 - F_\theta(\textbf{u}, \textbf{v})))
\end{align*}

\subsection*{Feature Prediction from Masked Window}
It is observed that networks which try to reconstruct every bit of the input waste capacity on modeling low-level details~\cite{oord2018representation}. Instead, in this auxiliary task we ask the network to approximate summary statistics of a masked temporal segment within a signal. To generate the data, we randomly sample the segment length $s_l \sim \mathcal{U}(n_{low}, n_{high})$ and starting point $s_p \sim \mathcal{U}(0, l - s_l)$. From the selected subsequence, we extract $8$ basic features: \texttt{mean}, \texttt{standard deviation}, \texttt{maximum}, \texttt{minimum}, \texttt{median}, \texttt{kurtosis}, \texttt{skewness}, \texttt{number of peaks}; and then mask the segment with zeros. The multi-head network is trained with Huber loss $\mathcal{L}_{HL}$ to predict the statistics of the missing subsequence for all modalities at the same time:

\begin{align*} 
\mathcal{L}_{HL}=\begin{cases}
	\frac{1}{2} \times o^2, & \text{if $|o| \leq \delta$}\\
    \delta \times (|o| - \frac{\delta}{2}), & \text{otherwise $|o| > \delta$} 
 \end{cases}, \text{where $o = F_\theta(\textbf{u}, \textbf{v}) - y$}  
\end{align*}

\subsection*{Transformation Recognition}
The signal transformation recognition is presented in~\cite{saeed2019multi} as an auxiliary task, where it is posed as a set of binary classification problems solved with a multi-task network to determine whether a signal is a transformed version or not. Here, we simplify the problem formulation and treat the task as multi-class classification, to learn a network that can directly recognize the applied transformation on an input from one out of $K$ classes. The benefits of our formulation are that it does not require specifying weights for task-specific losses and the network can be efficiently optimized with categorical cross-entropy objective $\mathcal{L}_{NL}$. Another key difference is that we address the problem of learning from multimodal data as opposed to a unimodal signal. To produce task-specific data, we generate transformed versions of each instance utilizing eight transformation functions: \texttt{permutation} , \texttt{channel shuffle}, \texttt{timewarp}, \texttt{scale}, \texttt{noise}, \texttt{rotation}, \texttt{flip}, \texttt{negation}), and an identity operation while assigning the function type as the corresponding class. During network training, we feed a batch of data consisting of examples for all the classes (inclusive of originals) and optimize a separate loss function for each input signal.  

\subsection*{Temporal Shift Prediction}
This conceptually straight-forward task consists of estimating the number of steps by which the samples are circularly-shifted in their temporal dimension. We pose this problem such that it can be treated either as a classification or as a regression task. We define a range of shift intervals, depending on the input resolution. For instance, in the activity recognition task, the considered ranges are: $[(0, 5), (6, 10), (11, 20), (21, 50), (51, 100), (101, 200), (201, \\ 300)]$. For producing shifted inputs, we first select a pair at random from the defined ranges, and second we sample a shifting factor within the defined boundary of the selected range. Last, we temporally shift the values of an input segment with the sampled factor. The network can be trained to predict either the range index (treating each entry as a class, with $7$ classes in total) or regress the factor. In our experiments, we notice that solving it as a regression problem results in better generalization on the end-task. Thus, the network is trained by minimizing mean-square error loss $\mathcal{L}_{MSE}$ for each sensing modality:          

\begin{align*} 
\mathcal{L}_{MSE} = \| F_\theta(\textbf{u},\textbf{v}) - y \|
\end{align*}

\subsection*{Modality Denoising}
This task's objective is to decompose a signal for obtaining a clean target through input reconstruction, i.e. isolating the mixed noise. It is similar in spirit to source separation in audio~\cite{luo2019convtasnet, zeghidour2020wavesplit} and to a denoising autoencoder, but without random noise~\cite{vincent2010stacked}. The fundamental intuition here is that if the network is tasked to reconstruct the original input from corrupted or mixed modality signals, then it forces the network to identify core signal characteristics while learning usable representations in the process. In our case, instead of mixing arbitrary noise, we exploit the availability of multisensor data to generate instances that might be of sufficient difficulty for the network to denoise. Specifically, we utilize a \texttt{weighted blending} operation $\mathbf{u} \times (1 - \mu) + \mathbf{v} \times \mu$ to mix instances of different modalities, i.e. we produce samples through combining the clean instances of accelerometer with gyroscope and vice versa while keeping the original samples as additional data points. The encoder-decoder network is trained end-to-end to minimize the mean-square error loss $\mathcal{L}_{MSE}$ between ground truth and corrupted input pairs.     

\subsection*{Odd Segment Recognition}
The goal of odd segment recognition is to identify the unrelated subsegment that does not belong to the input under consideration, where the rest of the sequences are in the correct order. The high-level idea behind the task is that if the network can spot artifacts in the signal, it should then also learn about useful input features. There are multiple ways to generate examples with odd subsegments; we approach it as an input consisting of an irregular segment of fixed length $s_o$ that is selected randomly from a different input modality. To generate proxy task examples, we begin with splitting an instance into equal-length sequences (e.g. of length $100$). Then, $2$ sequences from different modalities are randomly selected, that are either directly swapped or blended before applying a substitution operation. The index of the interchanged slices is used as the class, where valid inputs are assigned a distinct class. The network is asked to predict an index $id$ (out of 4 classes) of the odd sequence in each input modality. For this task, we minimize a categorical cross-entropy loss $\mathcal{L}_{NL}$ to train a multi-head network.

\subsection*{Metric Learning with Triplet Loss}
As we are interested in learning from multisensor data, we take advantage of multiple input modalities to formulate a metric learning objective. For this purpose, we utilize a symmetric triplet loss~\cite{zhang2016tracking}, which encourages the representations of similar inputs, while the representations of dissimilar inputs to be further apart. To optimize the specified loss, we need to generate input triplets consisting of an anchor, which can be an original instance, a positive sample that should be related (i.e. provides a complementary view of the input) to the anchor, and a negative sample which must be entirely different from the former pair. The loss then minimizes the distance between the anchor and the positive samples, while maximizing the distance of the negative samples from the anchor and the positive samples. For metric learning under this formulation, we generate the examples as follows: the actual instances are treated as anchors, and positive instances are generated by applying selected transformations at random on each anchor; whereas the negative instances are sampled from a different modality (i.e. for accelerometer, we treat samples from gyroscope as negatives). We then optimize $F_\theta$ with triplet loss $\mathcal{L}_{TL}$ to produce a smaller distance on associated samples and a more considerable distance on unrelated ones: 

\begin{align*} 
\mathcal{L}_{TL} = \max [0, \ D(z_{a}, z_{p}) - \frac{1}{2} \times (D(z_{a}, z_{n}) + D(z_{p}, z_{n})) + \alpha],
\end{align*}

\noindent where $z_{a}$, $z_{p}$, $z_{n}$ are the embeddings of anchor, positive and negative samples respectively, $\alpha$ represents the distance margin, and $D$ denotes squared-euclidean distance. 

\begin{algorithm}[htbp]
\caption{Sense and Learn}
\label{alg:sal}
\KwIn{Multisensory unlabeled data $\mathcal{D}_{U}$ and labeled data $\mathcal{D}_{L}$, auxiliary task $A_{t}$, number of iterations $I$, batch size $B$, L$2$ regularization rate $\beta$}
\KwOut{Self-supervised pre-trained network $F$}

initialize a representation learning network $F$ with parameters $\theta_{F}$\\
initialize a linear classifier $C$ with parameter $\theta_{C}$ for a down-stream task\\
initialize self-labeling data generation procedure $G_{T}$ based on task $A_{t}$\\
initialize proxy-task and end-task loss functions $\mathcal{L}_{T}$ and $\mathcal{L}_{E}$, respectively\\

\For{iteration $i$ $\in$ $\{$ $1$, \ $\ldots$, \ $I$ $\}$ }
{
    Randomly sample a mini-batch of $B$ instances from $\mathcal{D}_{U}$ as $\{x_1, x_2, \ldots, x_b\}$ \\
    Generate labeled (self-supervised) samples $\{$$(x$, $y)_1$, $(x$, $y)_2$, $\ldots$, and $(x$, $y)_b$$\}$ with $G_{T}$\\ 
    Update $\theta_F$ by descending along its gradient \\
    $\nabla_{\theta_{F}} \Big[\frac{1}{b} \sum_{i=1}^{B} \mathcal{L}_{T}(F_{\theta}(x_i), y_i) + \beta \left\lVert \theta\right\rVert^2 \Big]$
}

\For{iteration $i$ $\in$ $\{$ $1$, \ $\ldots$, \ $I$ $\}$ }
{
    Randomly sample a mini-batch of $B$ labeled instances from $\mathcal{D}_{L}$ as $\{$$(x$, $y)_1$, $(x$, $y)_2$, $\ldots$, and $(x$, $y)_b$$\}$ \\
    Extract latent embeddings $\textbf{z}$ from encoder $e$ within $F_{\theta}(\textbf{x})$\\
    Update $\theta_C$ by descending along its gradient \\
    $\nabla_{\theta_{c}} \Big[\frac{1}{b} \sum_{i=1}^{B} \mathcal{L}_{E}(C_{\theta}(z_i), y_i) \Big]$
}

We use Adam optimizer~\cite{kingma2014adam} for computing gradient-based parameter updates in all the experiments.
\end{algorithm}

\begin{figure}[t]
\centering
\includegraphics[width=11cm]{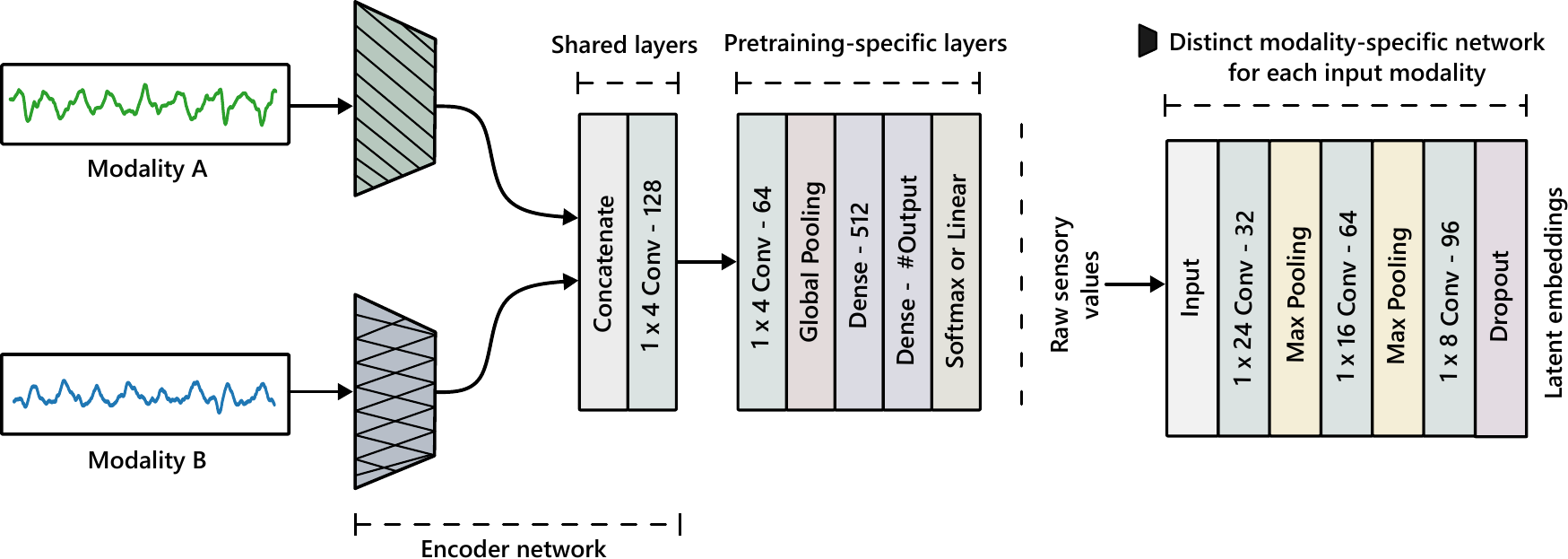}
\caption{A multistream neural network architecture for learning representations from multiple sensory inputs. A distinct stream (with an identical architecture) is used for each modality, as depicted on the right.}
\label{fig:architecture}
\end{figure}

\subsection{Network Architecture Design}
We implement the learning network $F_{\theta}$ as a multi-stream temporal convolutional model (TCN). The part of the motivation to use TCN came from~\cite{bai2018empirical} where it has been shown that convolutional networks perform remarkably well on sequence modeling tasks. Likewise, they have a low footprint for training and inference as compared to other methods and can be pruned easily to further compress the network~\cite{molchanov2016pruning}. The subnetworks share the same architecture, which is followed by a modality-agnostic network that fuses and learns a shared representation from the multimodal input. Jointly, we refer to these modules as encoder $e$, which is embedded within $F_{\theta}$. Importantly, we add an extra block connected to $e$, which is discarded after self-supervised pre-training. The intuition behind this strategy is that the model's last layers capture features that are primarily task-specific and do not generalize well on the end-task of interest. Therefore, the additional layers allow the base encoder to capture more generic features, while solving the auxiliary tasks. 

Figure~\ref{fig:architecture} illustrates the architecture design by precisely highlighting these main building blocks. The modality-specific encoder consists of three $1$D convolutional layers with $32$, $64$, and $96$ feature maps and a kernel size of $24$, $16$, and $8$, respectively. The max-pooling layer, with a pooling size of $4$ and a stride of $2$, is added after the initial convolutional layers. A dropout is used with a rate of $0.1$ at the end of the block. The shared encoder consists of a single convolutional layer with $128$ feature maps and a kernel size of $4$, which takes concatenated features as input. The supplementary layers in the pre-training block consist of a convolutional layer with $64$ feature maps and a kernel size of $4$ and a dense layer having $512$ hidden units. Importantly, a separate output layer is used for each input modality for all the surrogate tasks except `sensor blend,' which, based on its formulation, does not require this. Likewise, we use global pooling as the last layer in the representation learning network that aggregates discriminative features. L$2$ regularization with a rate of $0.0001$ is applied to the weights of all the layers to avoid overfitting. Moreover, we employ SELU as non-linearity except on the output layer; the network is trained with a learning rate of $0.0001$ for a maximum of $30$ epochs unless stated otherwise.     

We utilize a fixed network architecture for all the considered tasks (both auxiliary and down-stream), the intuition behind this choice being threefold. Firstly, we want to minimize the architectural differences to discover the true potential of self-supervision, i.e. it can be used with minimal effort on architecture tuning to extract semantic representations across diverse datasets. Secondly, our aim is to show that self-supervision has a huge prospect to be utilized for on-device learning. Having a smaller architecture and given the annotation-free nature of the proposed approach opens several exciting avenues in learning and inference with devices having limited processing capabilities. However, the further investigation of this is beyond the scope of our work and we leave it for future work. Lastly, our multi-modal architectural specification provides the flexibility to incorporate other modalities effortlessly. Furthermore, we highlight that in this work our focus is on individual task proposal and evaluation, but the framework can be used for jointly solving proxy tasks (i.e. in multi-task learning setting) as they share the same architecture, but differ fundamentally in terms of the loss function being optimized.

The high-level description of the learning procedure is summarized in Algorithm~\ref{alg:sal}. Given an unlabeled data $\mathcal{D}_{U}$ and a specified auxiliary task $A_{t}$, we optimize $F_{\theta}$ with task-specific data that is generated on-the-fly, as described in the preceding section. Once pre-training converges, the layers specific to self-supervised learning are discarded, and the encoder $e$ is saved. Then, the second round of training on a down-stream task of interest begins with labeled data $\mathcal{D}_{L}$. Depending on the evaluation criteria, the following can be done: a) the network is either kept frozen and used as a generic feature extractor for learning a linear classifier\footnote{logistic regression}, b) the modality-agnostic encoder $e_{s}$ is fine-tuned during learning an end-task, or c) the self-supervised network is used as initialization for rapidly solving the final-task, e.g., fine-tuning a model with little labeled data. The \textit{encoder} network shown in Figure~\ref{fig:architecture} represents the module that is kept frozen, while depending on the learning setting the \textit{shared} layers are further fine-tuned.

\section{Experiments}
\label{experiments}
We perform a comprehensive evaluation of our framework on four different application domains: a) activity recognition, b) sleep-stage scoring, c) stress detection, and d) WiFi sensing. For every area, we train the self-supervised networks with each proposed task and determine the quality of the learned representation with either a linear classifier or by fine-tuning with few labeled instances. Furthermore, we also examine the knowledge transferability between related datasets. In the following, we describe the utilized datasets, pre-processing steps, and assessment strategy, including the baselines.    

\subsection{Datasets}
\label{sec:datasets}
We assess the performance of \textit{Sense and Learn} on $8$ publicly available multisensor datasets from diverse domains. The brief description of each utilized data source is provided below, with Table~\ref{tab:dataset} summarizing their major characteristics. 

\begin{table}[!htbp]
\centering
\caption{Key characteristics of the datasets used in the experiements. The relative class distribution of each dataset is given in Figure~\ref{fig:cd} of appendix~\ref{appendix:class_distribution}.}
\label{tab:dataset}
\small
\begin{tabular}{ccccc}
\hline
\textbf{Dataset} & \textbf{\#Subjects} & \textbf{\#Classes} & \textbf{Task} & \textbf{Inputs} \\ \hline
HHAR & 9 & 6 & \multirow{5}{*}{\begin{tabular}[c]{@{}c@{}}Activity/Context\\ Recognition\end{tabular}} & \multirow{5}{*}{\begin{tabular}[c]{@{}c@{}}Accelerometer \\ \&\\ Gyroscope\end{tabular}} \\
MobiAct & 66 & 11 &  &  \\
MotionSense & 24 & 6 &  &  \\
UCI HAR & 30 & 6 &  &  \\
HAPT & 30 & 12 &  &  \\ \hline
Sleep-EDF & 20 & 5 & Sleep Stage Scoring & EEG \& EOG \\ \hline
MIT DriverDb & 17 & 2 & Stress Detection & \begin{tabular}[c]{@{}c@{}}Heart Rate \& \\ Skin Conductance\end{tabular} \\ \hline
WiFi CSI & 6 & 7 & \begin{tabular}[c]{@{}c@{}}Activity (Behavior)\\ Recognition\end{tabular} & CSI Amplitude \\ \hline
\end{tabular}
\end{table}

\subsubsection*{Activity Recognition}
For smartphone-based human activity recognition, we select $5$ datasets containing accelerometer and gyroscope signals, namely: HHAR, MobiAct, UCI HAR, MotionSense, and HAPT. The Heterogeneity Human Activity Recognition (HHAR) dataset~\cite{stisen2015smart} is collected from $9$ participants, each performing $6$ basic activities (i.e. sitting, standing, walking, stairs-up, stairs-down and biking) for $5$ minutes. A broad range of devices is used for the systematic analysis of sensor, device, and workload-specific heterogeneities across manufacturers. Specifically, each user carried $8$ smartphones on different body locations that were selected from a pool of $36$ devices of different models and brands. Likewise, the sampling rate differs considerably across phones with values ranging between 50Hz-200Hz. The MotionSense dataset~\cite{malekzadeh2018protecting} is recorded with the aim of inferring personal attributes, such as physical and demographics, in addition to the activities. The iPhone$6$s is placed in the users' front pocket during the collection phase, while they performed $15$ trials of $6$ activities in the same experimental setting. In total, $24$ subjects of varying height, weight, age and gender performed the following 6 activities: walking, jogging, sitting, standing, downstairs and upstairs. We use this data only for the detection of activities without concerning with the identification of other attributes. UCI HAR~\cite{anguita2013public} comprises data obtained from $30$ subjects with waist-mounted Samsung Galaxy S$2$ devices sampling at $50$Hz. Each participant completed $6$ activities of daily living (i.e. standing, sitting, lying down, walking, downstairs and upstairs) during $2$ trials with a $5$ seconds resting condition in-between. The MobiAct~\cite{Vavoulas2014TheMD} contains inertial sensors data collected from $66$ participants with Samsung Galaxy S$3$ phones through more than $3200$ trails. The subjects freely placed the device in their trouser's pocket to mimic real-life phone usage and placement. We utilize the data of $61$ subjects for whom data of any of the following $11$ activity classes is available: walking, jogging, jumping, upstairs, downstairs, sitting, stand to sit, sit to stand, sitting on a chair, car step-in and car step-out. The Human Activities and Postural Transitions (HAPT) dataset~\cite{reyes2016transition} is collected from a group of $30$ volunteers with Samsung Galaxy S$2$ devices sampling at $50$Hz. The phone was mounted on the waist of each subject who completed $3$ dynamic activities (walking,  upstairs, downstairs), $3$ static posture activities (lying, sitting, standing), and $6$ postural transitions (sit-to-lie, lie-to-sit, stand-to-sit, sit-to-stand, stand-to-lie, and lie-to-stand); resulting in $12$ classes.   

\subsubsection*{Sleep Stage Scoring}
We use the PhysioNet Sleep-EDF\footnote{version 1} dataset~\cite{kemp2000analysis, goldberger2000physiobank} consisting of $61$ polysomnograms (PSGs) from $20$ subjects. It is comprised of participants from $2$ different studies: a) effect of age on sleep and b) Temazepam effect on sleep. We use the $2$ whole-night PSGs sleep recording sampled at $100$Hz from the former study. Each record contains $2$ electroencephalogram (EEG) signals from Fpz-Cz and Pz-Oz electrode locations, electrooculography (EOG), electromyography (EMG) and event markers. Some instances also have oro-nasal respiration and body temperature. The hypnograms (30-seconds1 epochs) were manually annotated by sleep expert with one of the $8$ sleep classes (Wake, N$1$, N$2$, N$3$, N$4$, Rapid Eye Movement, Movement, Unknown), based on the R\&K standard. We utilize EEG (Fpz-Cz) and EOG signals in our evaluation. Following previous work~\cite{supratak2017deepsleepnet}, we merged N$3$ and N$4$ into a single class N$3$ and discarded Movement and unscored samples, to have $5$ sleep stages.

\subsubsection*{Stress Detection}
For physiological stress recognition, we utilize the MIT DriverDb dataset~\cite{healey2005detecting, goldberger2000physiobank}, which is collected during a real-world driving experiment in a city, on a highway and in a resting condition. The publicly-available version on PhysioNet consists of $17$ drives out of $24$, each lasted between $1$-$1.5$ hours. The following physiological signals are recorded: EMG, electrocardiography (ECG), galvanic skin response (GSR) from hand and foot, heart rate (HR; derived from ECG), and breathing rate. The signals were originally sampled at different rates but downsampled to $15.5$Hz. The `marker' signal provided in the dataset is used to derive the binary ground truth, indicating a change-of-drive (i.e. resting, city or highway driving), which is found to be correlated with distress level through post-driving video analysis by experts~\cite{healey2005detecting}. We use the following $10$ drives $04$, $05$, $06$, $07$, $08$, $09$, $10$, $11$, $12$ and $16$ in our experiments, which have HR and GSR (from hand), given collection of other signals in real-life is quite problematic.

\subsubsection*{WiFi Sensing}
Device-free context recognition with WiFi is an emerging area of research. To show the robustness of our self-supervised methods on this task, particularly on a unimodal signal, we utilize the WiFi channel state information (CSI) dataset~\cite{yousefi2017survey} for activity recognition. This dataset is collected in a controlled office environment, where the transmitting (router) and receiving (Intel 5300 NIC) devices were 3m apart, and the channel state information (CSI) was recorded at $1$kHz. The $6$ subjects performed $20$ trials for each of the following $7$ activities: lying down, falling, walking, running, sitting down, standing up and picking something up. The ground truth was obtained from videos recorded during the data collection process, and CSI amplitude is used for learning a model.

\subsection{Pre-processing and Assessment Strategy}
To prepare the data for sequence modeling with a temporal convolutional network, we utilize a sliding window approach to segment the signals into fixed-sized inputs. In the case of the activity recognition task, we choose a window size of $400$ samples with a $50\%$ overlap, except for the HAPT dataset where a segment size of $200$ samples is used, due to the short duration of posture-transition activities. We found these windows sizes to be optimal based on earlier experiments, as each activity dataset has a different sampling rate. We did not perform resampling as the sampling rate differences among phones does not vary significantly and $1$D convolutional layers with wide kernel sizes learn to adapt to the specific characteristics of the input signal. However, if the sampling rate varies considerably it might be essential to do resampling. For Sleep-EDF, we applied minimal pre-processing based on existing work~\cite{supratak2017deepsleepnet} to formulate the problem as a $5$-stage sleep classification and used the $30$ seconds epochs as model input. In the WiFi sensing task, we process the input same as the original work that open-sourced the data and utilize a downsampled CSI signal of $500$Hz as~\cite{yousefi2017survey}, which corresponds to an input window of $1$ second. The heart rate and skin conductance signals from MIT DriverDb are processed to remove artifacts  and these signals are mean normalized using the `mean' and `standard deviation' calculated from the baseline (or resting phase) of the data collection following~\cite{saeed2017personalized} for each subject. We use a window size of $30$ seconds with $50\%$ overlap to generate input segments for the model. We randomly split the datasets based on subjects into train and test sets withholding $70\%$ users for training and the rest $30\%$ for testing. We further divide the training set to obtain a validation set of size $20\%$, which is used for hyper-parameter tuning and early stopping. Most importantly, we also perform $5$-fold cross-validation for thorough performance analysis whenever it is applicable. Furthermore, we z-normalize the samples with mean and standard deviation calculated from the training set. For self-supervision, we pre-train the models using only the training set, including for the transfer learning experiments. The self-labeled examples are generated for each task on-the-fly during the learning phase, as defined earlier in Section~\ref{sec:sslt}. 

For each recognition problem, we treat a fully-supervised model directly trained (in an end-to-end manner) with the annotated data of an end-task as a `baseline.' Likewise, we compare self-supervised tasks against pre-training with a standard autoencoder. As explained earlier, we assess the quality of the self-supervised representation (including in the transfer-learning setting) through training a linear classifier or fine-tuning the last convolutional layer of the encoder on the downstream tasks. For learning in the low-data regime, we use a self-supervised network as initialization to quickly learn a model with few labeled examples. In all the cases, we assess the network performance with a weighted version of F-score and Cohen's kappa (see appendix~\ref{appendix:kappa_results});  as these metrics are robust to unbalanced class distributions while being sensitive to misclassifications. 

\subsection{Results and Discussion}

\begin{table}[t]
  \caption{Performance evaluation (weighted F-score) of self-supervised representations with a linear classifier. The unsupervised pre-trained networks achieve competitive performance with the fully-supervised networks. In WiFi-CSI sub-table, the entries with hyphen indicate auxiliary tasks that cannot be applied to unimodal signals. See Table~\ref{tab:kappa_linear} in appendix~\ref{appendix:kappa_results} for kappa scores.}\label{tab:linear}
  \centering
  \subfloat{
    \small
    \centering
    \begin{tabular}{ccccc}
    Method & \textbf{HHAR} & \textbf{MobiAct} & \textbf{MotionSense} & \textbf{UCI HAR} \\ \hline
    Fully Supervised & 0.794$\pm$0.014 & 0.934$\pm$0.005 & 0.952$\pm$0.007 & 0.962$\pm$0.006 \\
    Random Init. & 0.218$\pm$0.062 & 0.383$\pm$0.109 & 0.246$\pm$0.090 & 0.221$\pm$0.079 \\
    Autoencoder & 0.777$\pm$0.003 & 0.726$\pm$0.001 & 0.675$\pm$0.019 & 0.782$\pm$0.042 \\ \hline
    Sensor Blend & 0.823$\pm$0.006 & \cellcolor[gray]{0.93}0.912$\pm$0.001 & \cellcolor[gray]{0.93}0.911$\pm$0.009 & 0.902$\pm$0.010 \\
    Fusion Magnitude & \cellcolor[gray]{0.93}0.848$\pm$0.005 & 0.905$\pm$0.001 & \cellcolor[gray]{0.93} 0.925$\pm$0.011 & 0.895$\pm$0.010 \\
    Feature Prediction & 0.817$\pm$0.005 & 0.902$\pm$0.001 & 0.849$\pm$0.010 & 0.899$\pm$0.010 \\
    Transformations & \cellcolor[gray]{0.93} 0.854$\pm$0.005 & \cellcolor[gray]{0.93}0.911$\pm$0.002 & 0.869$\pm$0.013 & \cellcolor[gray]{0.93}0.906$\pm$0.011 \\
    Temporal Shift & 0.834$\pm$0.008 & 0.909$\pm$0.003 & 0.851$\pm$0.016 & 0.747$\pm$0.027 \\
    Modality Denoise. & 0.807$\pm$0.006 & 0.817$\pm$0.004 & 0.675$\pm$0.019 & 0.798$\pm$0.035 \\
    Odd Segment & 0.835$\pm$0.006 & 0.901$\pm$0.001 & 0.869$\pm$0.012 & 0.888$\pm$0.010 \\
    Tripet Loss & 0.773$\pm$0.005 & 0.841$\pm$0.002 & 0.910$\pm$0.008 & \cellcolor[gray]{0.93}0.905$\pm$0.011 \\ \hline
    \end{tabular}
  } \hspace{0.01cm}
  \subfloat{
    \small
    \centering
    \begin{tabular}{ccccc}
    Method & \textbf{HAPT} & \textbf{Sleep-EDF} & \textbf{MIT DriverDb} & \textbf{WiFi CSI} \\ \hline
    Fully Supervised & 0.899$\pm$0.009 & 0.825$\pm$0.005 & 0.824$\pm$0.029 & 0.964$\pm$0.007 \\
    Random Init. & 0.119$\pm$0.041 & 0.149$\pm$0.127 & 0.321$\pm$0.198 & 0.153$\pm$0.04 \\ 
    Autoencoder & 0.669$\pm$0.003 & 0.679$\pm$0.012 & 0.876$\pm$0.002 & 0.767$\pm$0.005 \\ \hline
    Sensor Blend & 0.818$\pm$0.006 & 0.779$\pm$0.004 & 0.890$\pm$0.002 & - \\
    Fusion Magnitude & 0.815$\pm$0.004 & \cellcolor[gray]{0.93}0.782$\pm$0.006 & 0.892$\pm$0.004 & - \\
    Feature Prediction & \cellcolor[gray]{0.93}0.822$\pm$0.002 & 0.671$\pm$0.022 & 0.866$\pm$0.000 & \cellcolor[gray]{0.93}0.837$\pm$0.005 \\
    Transformations & \cellcolor[gray]{0.93}0.841$\pm$0.003 & 0.778$\pm$0.006 & \cellcolor[gray]{0.93}0.908$\pm$0.001 & 0.768$\pm$0.007 \\
    Temporal Shift & 0.782$\pm$0.004 & 0.707$\pm$0.012 & 0.883$\pm$0.005 & 0.731$\pm$0.011 \\
    Modality Denoise. & 0.738$\pm$0.002 & \cellcolor[gray]{0.93}0.784$\pm$0.002 & 0.902$\pm$0.001 & - \\
    Odd Segment & 0.790$\pm$0.003 & 0.772$\pm$0.003 & \cellcolor[gray]{0.93}0.885$\pm$0.002 & 0.774$\pm$0.008 \\
    Tripet Loss & 0.815$\pm$0.002 & 0.775$\pm$0.003 & 0.891$\pm$0.001 & 0.749$\pm$0.009 \\ \hline
    \end{tabular}
  }
\end{table}

\subsubsection*{Linear separability and effects of fine-tuning the shared encoder}
\label{subsec:linear}
For assessing the quality of the self-supervised embeddings, we conduct experiments with a linear classifier on the end-tasks. Linear separability is a standard way of measuring the power of self-supervised-learned features in the literature~\cite{oord2018representation, tagliasacchi2019self, gidaris2018unsupervised}, i.e. if the representations disentangle factors of variations in the input, then it becomes easier to solve subsequent tasks. Here, we train a linear classifier (i.e. logistic regression) $10$-times on top of a frozen network (pre-trained with self-supervision) using annotated data of the downstream task. Table~\ref{tab:linear} summarizes the results on eight benchmark datasets from four application domains. We compare the performance against a fully-supervised network that is trained in an end-to-end manner (directly with annotated data). We also consider unsupervised pre-training with a standard autoencoder to analyze the improvements of self-supervision. Likewise, a linear model is also trained with random features (i.e. from a randomly initialized frozen network) to estimate its learning capacity. On the activity recognition problem, the self-supervised features achieve very close results on multiple benchmarks to training an entire network with annotated instances. On the HHAR dataset, the transformation and fusion magnitude prediction tasks improve the F-score by $7$ points. On other datasets with a large number of classes, such as HAPT and MobiAct, our simple proxy tasks learn features that are generalizable to end-tasks. In the case of sleep stage scoring, linear layers trained with features from the modality denoising and the fusion magnitude tasks achieve a kappa of $0.70$, which is impressive given that the representations are learned from completely unlabeled data. Similarly, in a stress classification problem, the self-supervised networks outperform a fully-supervised model with a large margin. The transformations and modality denoising tasks achieve kappa scores of $0.80$ and $0.79$, respectively. We believe it is because pre-training results in generic features, whereas a model trained directly on the end-task suffers from overfitting. Lastly, we evaluate on the device-free sensing problem using the amplitude of WiFi CSI. Although we designed the auxiliary tasks for multisensorinput, we find a subset of these to be applicable for self-supervision with a unimodal input. We achieve good results with self-supervised features even though the dataset size is relatively small, and input is noisy, complex and high-dimensional. The linear layer trained on top of the feature-prediction task representations achieves an F-score of $83\%$ compared to the end-to-end training F-score of $96\%$.

\begin{table}[htbp]
   \caption{Improvement in recognition rate (weighted F-score) by fine-tuning the shared layers of the encoder while training on the end-task. We observe a significant increase in performance across datasets with self-supervised networks, either surpassing or achieving results on-par with the baseline. See Table~\ref{tab:kappa_ft} in appendix~\ref{appendix:kappa_results} for kappa scores.}\label{tab:ft}
   \centering
   \subfloat{
    \small
    \centering
    \begin{tabular}{ccccc}
    Method & \textbf{HHAR} & \textbf{MobiAct} & \textbf{MotionSense} & \textbf{UCI HAR} \\ \hline
    Fully Supervised & 0.794$\pm$0.014 & 0.934$\pm$0.005 & 0.952$\pm$0.007 & 0.961$\pm$0.008 \\
    Random Init. & 0.218$\pm$0.062 & 0.383$\pm$0.109 & 0.246$\pm$0.090 & 0.221$\pm$0.079 \\
    Autoencoder & 0.835$\pm$0.003 & 0.927$\pm$0.003 & 0.938$\pm$0.002 & 0.943$\pm$0.004 \\ \hline
    Sensor Blend & \cellcolor[gray]{0.93}0.841$\pm$0.009 & \cellcolor[gray]{0.93}0.943$\pm$0.004 & 0.937$\pm$0.004 & \cellcolor[gray]{0.93}0.956$\pm$0.003 \\
    Fusion Magnitude & 0.831$\pm$0.006 & 0.938$\pm$0.005 & 0.945$\pm$0.002 & 0.946$\pm$0.002 \\
    Feature Prediction & \cellcolor[gray]{0.93}0.840$\pm$0.007 & 0.937$\pm$0.002 & 0.951$\pm$0.003 & 0.943$\pm$0.003 \\
    Transformations & 0.828$\pm$0.006 & \cellcolor[gray]{0.93}0.946$\pm$0.004 & \cellcolor[gray]{0.93}0.951$\pm$0.005 & \cellcolor[gray]{0.93}0.954$\pm$0.006 \\
    Temporal Shift & 0.831$\pm$0.008 & 0.939$\pm$0.002 & 0.934$\pm$0.006 & 0.909$\pm$0.008 \\
    Modality Denoise. & 0.840$\pm$0.003 & 0.938$\pm$0.002 & 0.928$\pm$0.006 & 0.941$\pm$0.001 \\
    Odd Segment & 0.826$\pm$0.003 & 0.938$\pm$0.005 & 0.935$\pm$0.006 & 0.953$\pm$0.003 \\
    Tripet Loss & 0.835$\pm$0.013 & 0.912$\pm$0.006 & \cellcolor[gray]{0.93}0.955$\pm$0.003 & 0.950$\pm$0.002 \\ \hline
    \end{tabular} 
   } \hspace{0.02cm}
   \subfloat{
    \small
    \centering
    \begin{tabular}{ccccc}
    Method & \textbf{HAPT} & \textbf{Sleep-EDF} & \textbf{MIT DriverDb} & \textbf{WiFi CSI} \\ \hline
    Fully Supervised & 0.899$\pm$0.009 & 0.825$\pm$0.005 & 0.824$\pm$0.029 & 0.964$\pm$0.007 \\
    Random Init. & 0.119$\pm$0.041 & 0.149$\pm$0.127 & 0.321$\pm$0.198 & 0.153$\pm$0.048 \\
    Autoencoder & 0.881$\pm$0.002 & 0.805$\pm$0.008 & 0.877$\pm$0.002 & \cellcolor[gray]{0.93}0.898$\pm$0.025 \\ \hline
    Sensor Blend & 0.895$\pm$0.003 & 0.809$\pm$0.003 & 0.881$\pm$0.014 & - \\
    Fusion Magnitude & 0.898$\pm$0.002 & 0.813$\pm$0.003 & 0.882$\pm$0.011 & - \\
    Feature Prediction & 0.893$\pm$0.003 & 0.748$\pm$0.006 & 0.859$\pm$0.003 & 0.832$\pm$0.037 \\
    Transformations & \cellcolor[gray]{0.93}0.898$\pm$0.002 & \cellcolor[gray]{0.93}0.822$\pm$0.005 & \cellcolor[gray]{0.93}0.890$\pm$0.005 & 0.823$\pm$0.028 \\
    Temporal Shift & 0.876$\pm$0.007 & 0.779$\pm$0.005 & 0.883$\pm$0.005 & 0.736$\pm$0.063 \\
    Modality Denoise. & 0.885$\pm$0.003 & \cellcolor[gray]{0.93}0.819$\pm$0.002 & \cellcolor[gray]{0.93}0.889$\pm$0.001 & - \\
    Odd Segment & \cellcolor[gray]{0.93}0.899$\pm$0.003 & 0.804$\pm$0.003 & 0.853$\pm$0.023 & \cellcolor[gray]{0.93}0.860$\pm$0.030 \\
    Tripet Loss & 0.887$\pm$0.005 & 0.805$\pm$0.003 & 0.884$\pm$0.002 & 0.755$\pm$0.022 \\ \hline
    \end{tabular}
   } 
\end{table}

In Table~\ref{tab:ft}, we notice a substantial improvement on the downstream tasks if the last convolutional layer of the encoder (see Figure~\ref{fig:architecture}) is fine-tuned while training the linear classifier. Comparing with the results given in Table~\ref{tab:linear}, it can be seen that the recognition rate of the models improved significantly, achieving similar results as the fully-supervised baselines; while features learned by input reconstruction with an autoencoder scored low compared to our proposed surrogate tasks even after fine-tuning, except for the WiFi sensing task. On the MobiAct dataset, transformations and sensor blend tasks gain ~$2$ points improvement in kappa. Likewise, for MotionSense, HAPT and UCI HAR, we bridge the gap between fully-supervised and self-supervised models. Interestingly, fine-tuning did not help much with MIT DriverDb compared to training a linear classifier. These results agree with our intuition that training on an end-task directly in this case results in overfitting.   

In summary, the evaluation with a linear classifier trained on top of a pre-trained (self-supervised) feature extractor highlights that the representations learned with auxiliary tasks are broadly useful and better than autoencoding-based approaches. It also confirms our hypothesis that general-purpose representations can be learned directly from raw input without any strongly (task-specific) labeled data. It is important to note we did not aim to surpass fully-supervised approaches in this setting. Supervised methods will be better because they have direct access to task-specific labels, while self-supervised objectives train a network without any foresight of the end-task. It can also be seen from the results of fine-tuning the encoder, as presented in Table~\ref{tab:ft}, that the network performance matches the supervised methods or improves upon, when shared layers are further trained on the downstream tasks. Likewise, it might be possible to improve generalization of self-supervised models through pre-training on larger unlabeled datasets in a real-world setting.

\subsubsection*{Impact on learning in low-data regime}
\begin{figure}[htbp]
\centering
\subfloat[HHAR]{\includegraphics[width=6.4cm]{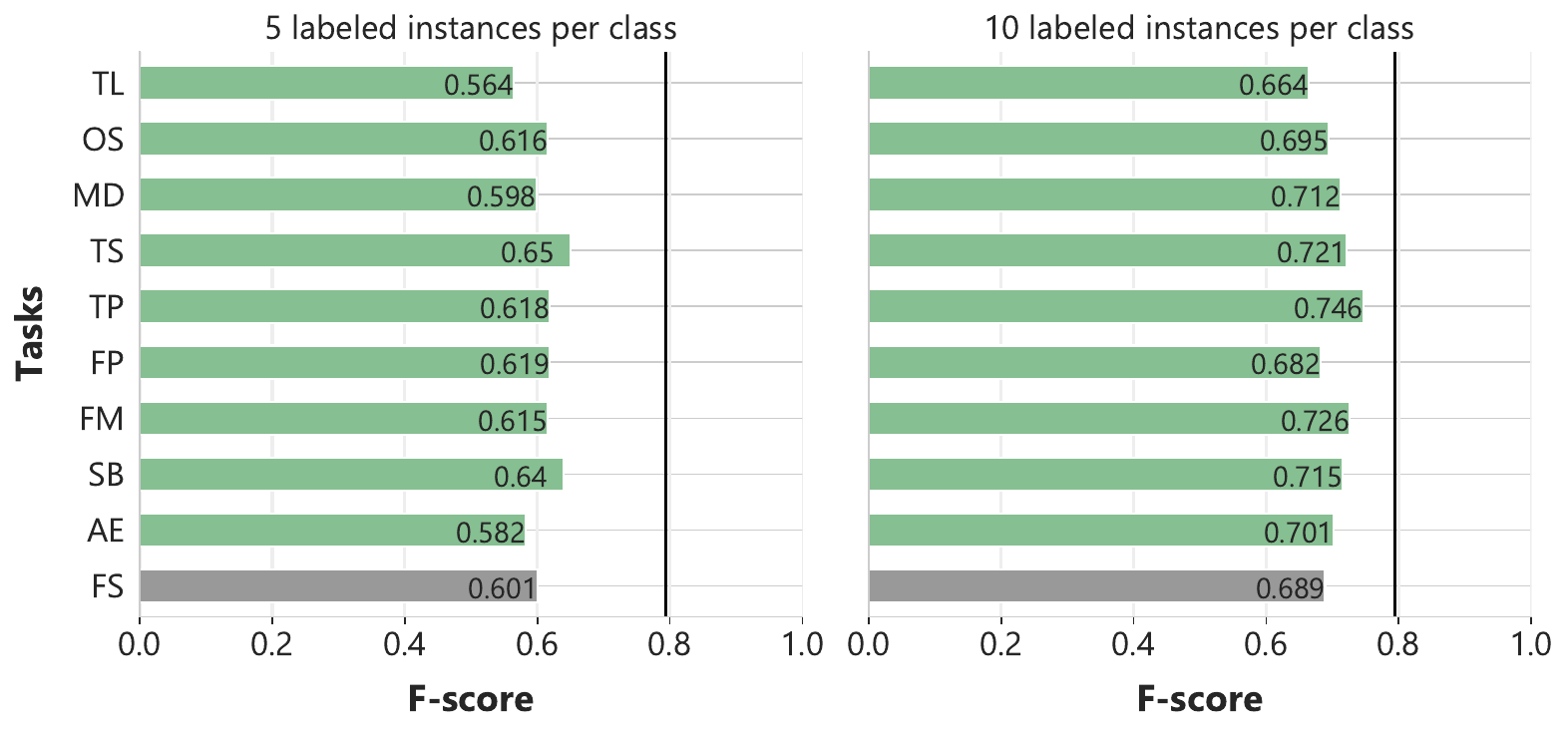}}  
\subfloat[MobiAct]{\includegraphics[width=6.4cm]{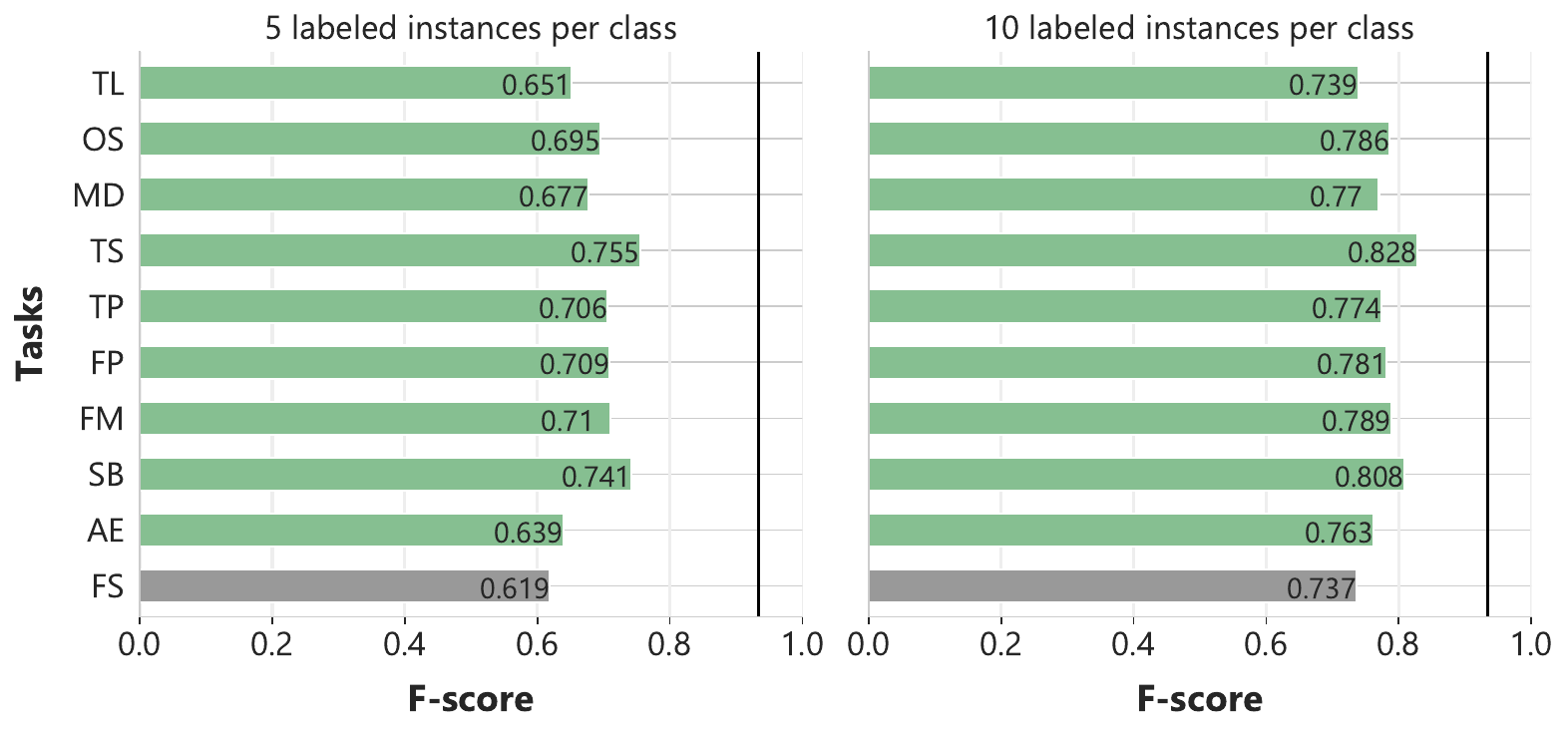}}\\
\subfloat[MotionSense]{\includegraphics[width=6.4cm]{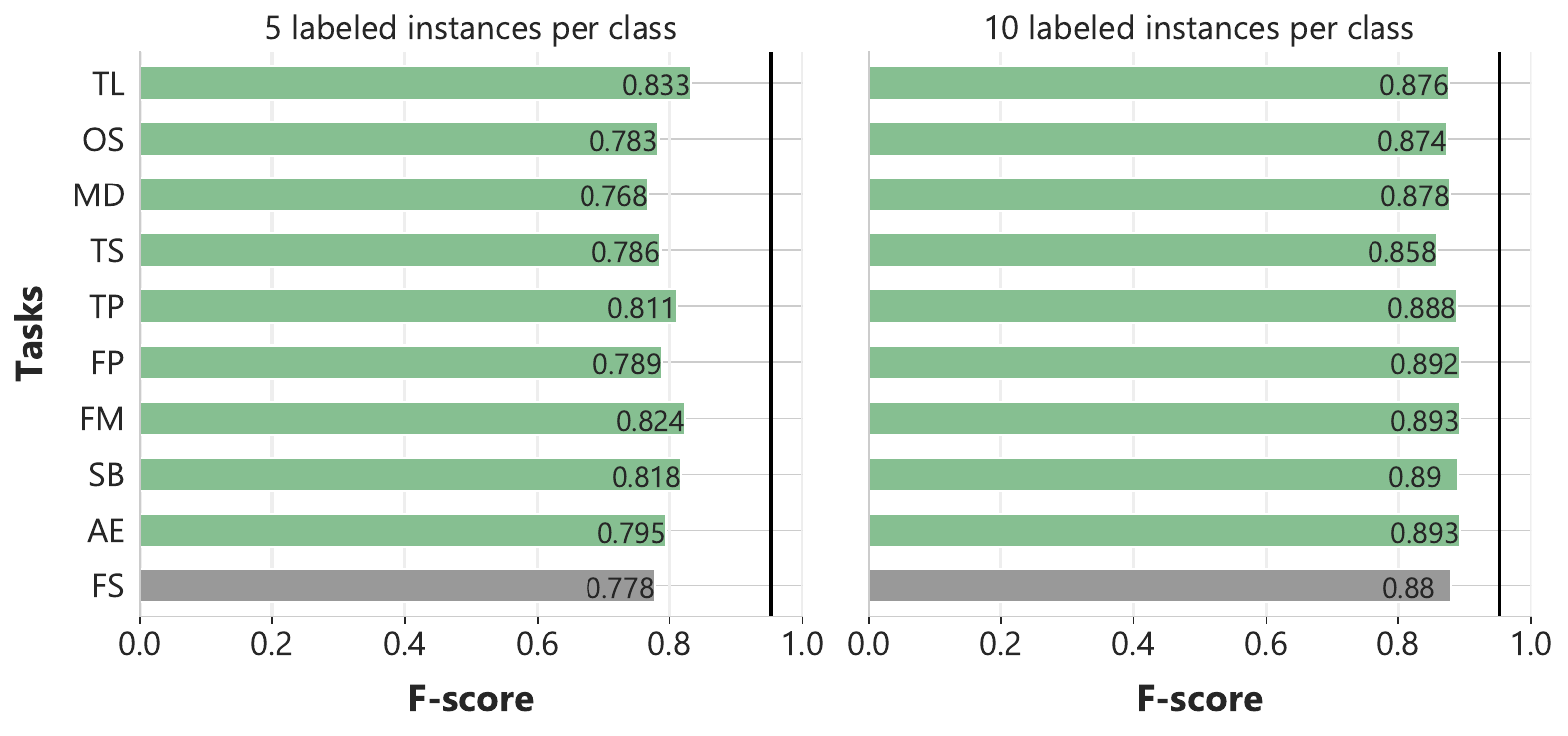}} 
\subfloat[UCI HAR]{\includegraphics[width=6.4cm]{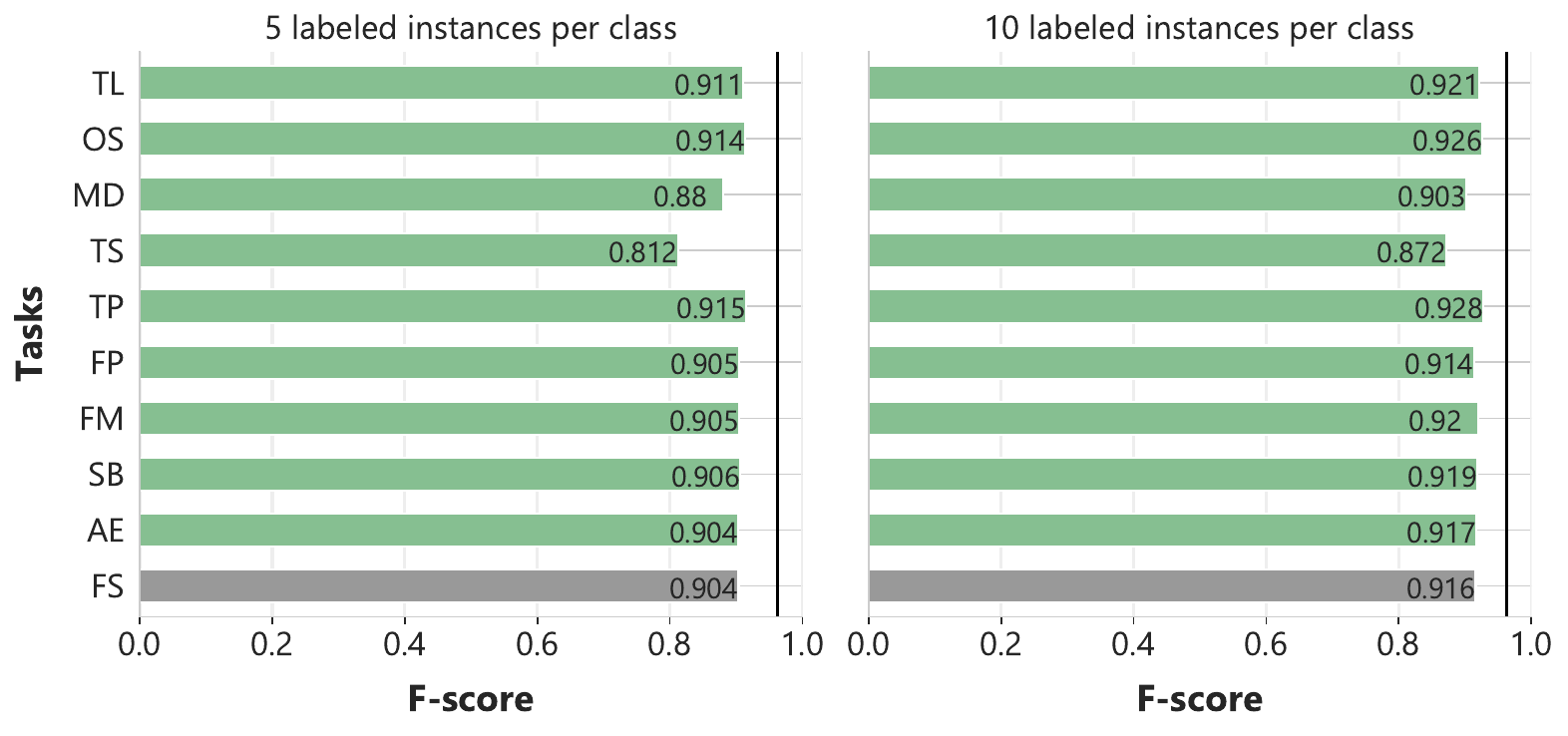}} \\
\subfloat[HAPT]{\includegraphics[width=6.4cm]{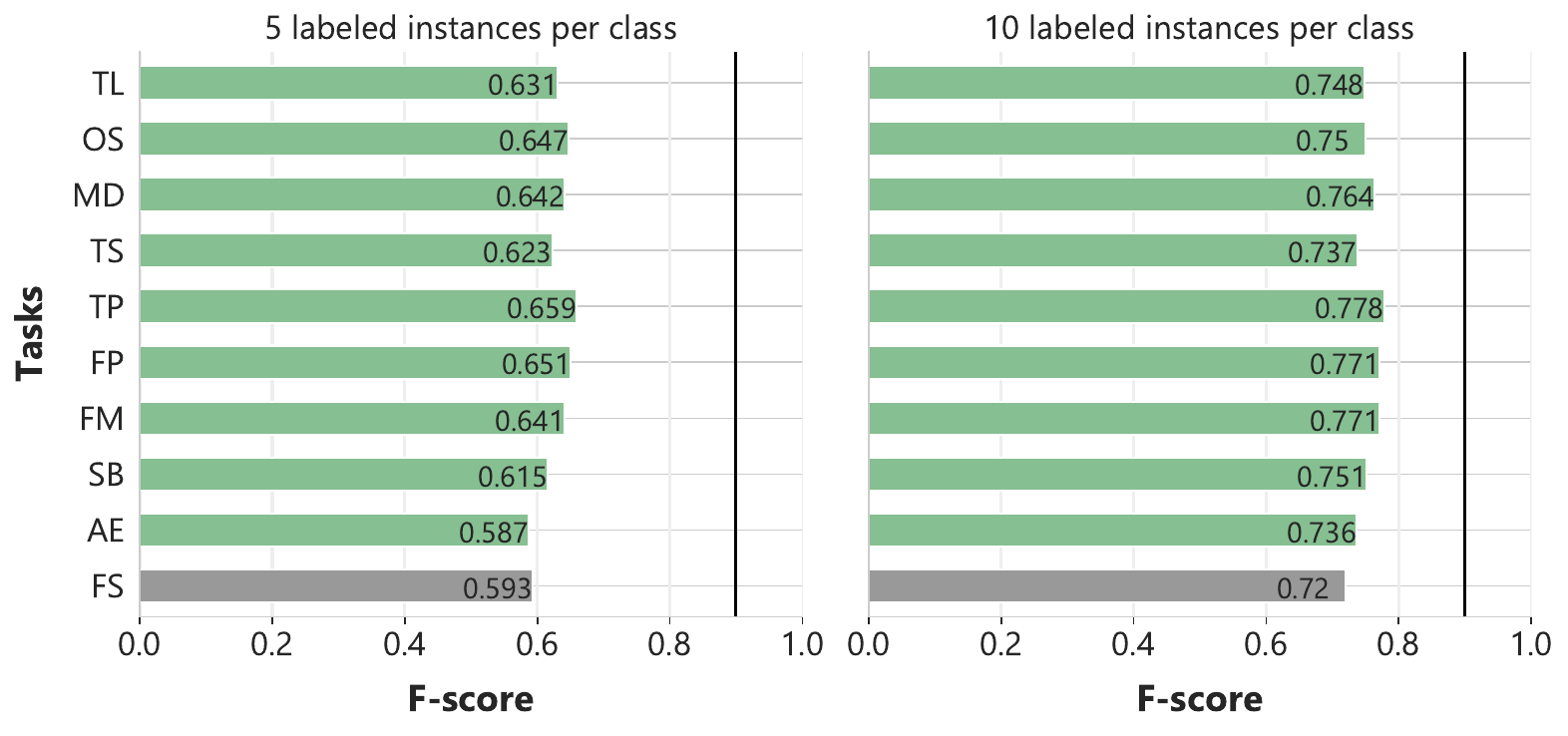}} 
\subfloat[Sleep-EDF]{\includegraphics[width=6.4cm]{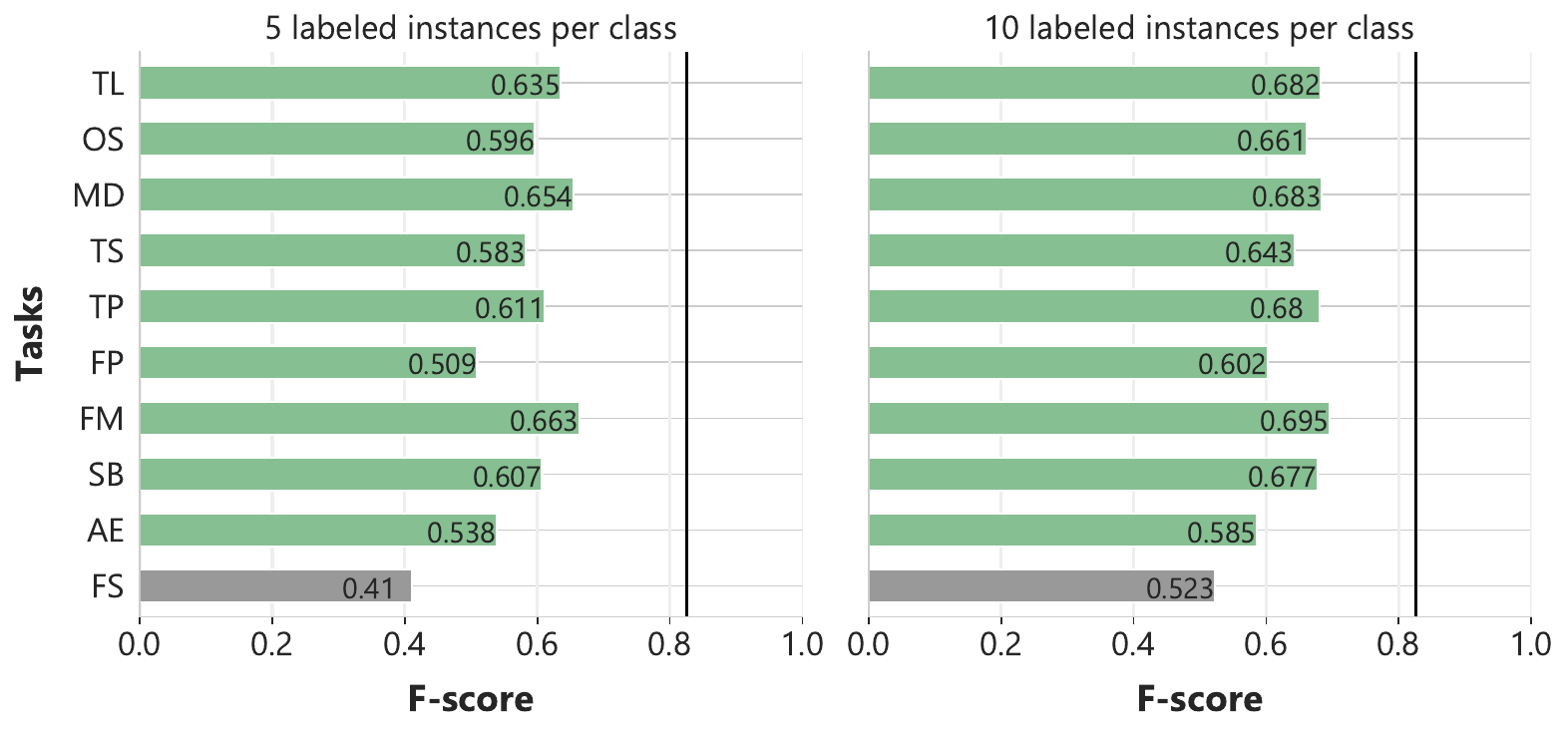}} \\
\subfloat[MIT DriverDb]{\includegraphics[width=6.4cm]{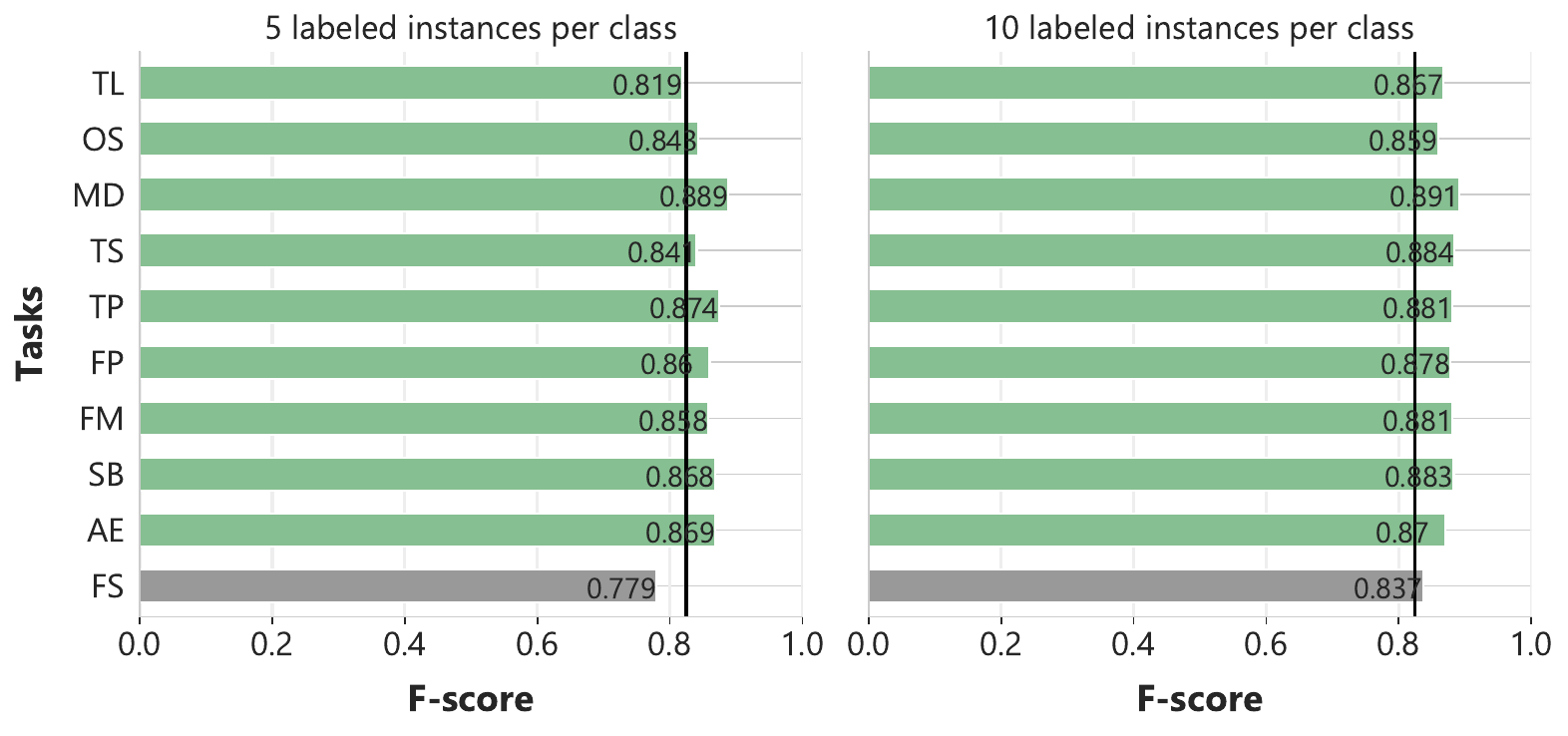}} 
\subfloat[WiFi CSI]{\includegraphics[width=6.4cm]{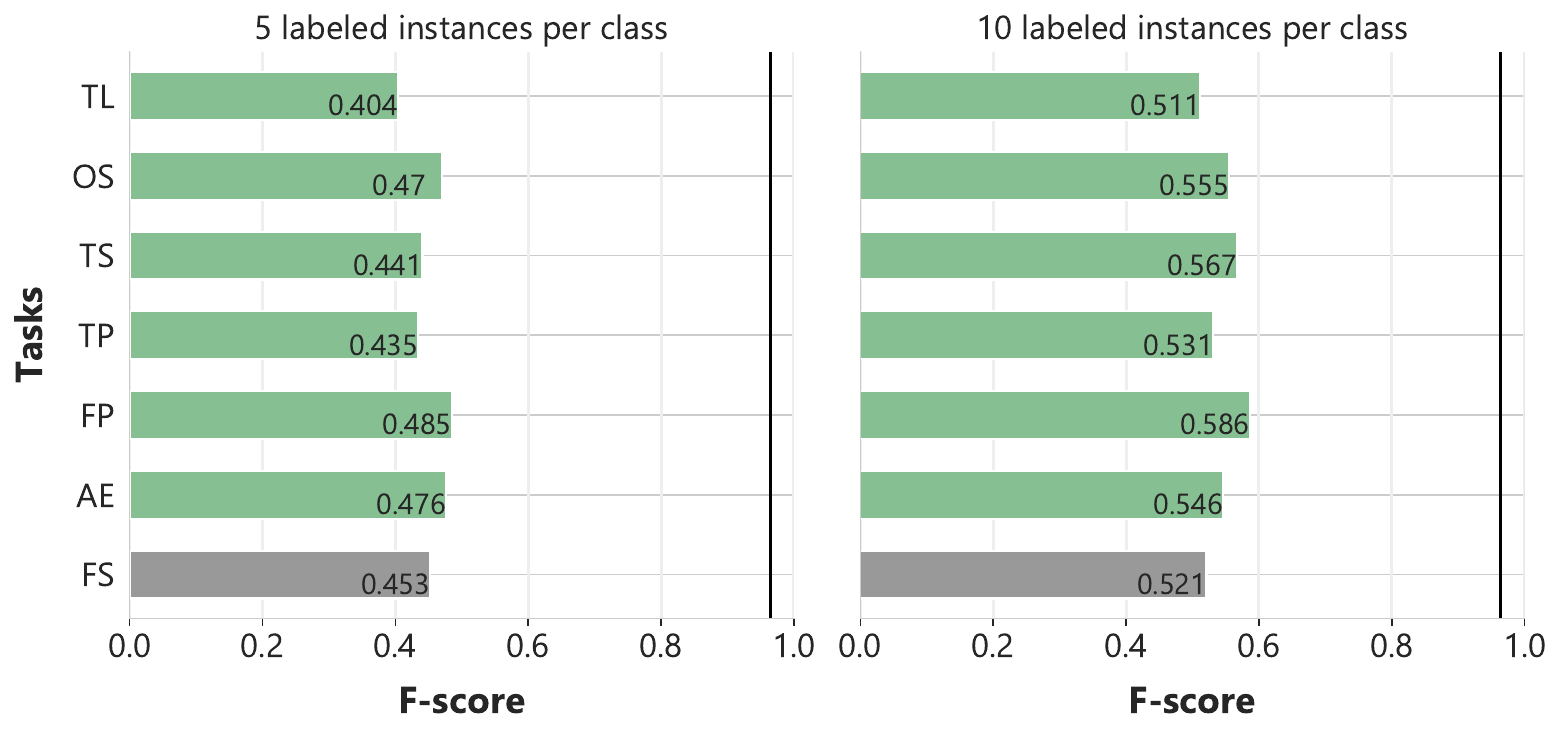}}\\
\subfloat{\includegraphics[width=6.4cm]{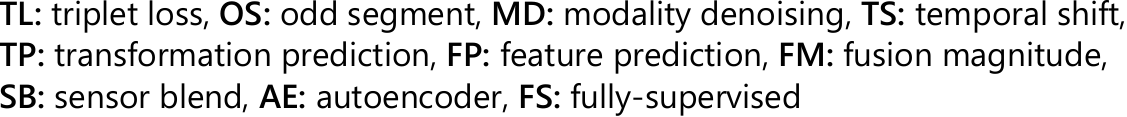}}
\caption{Contribution of self-supervised pre-training for improving end-task performance with few labeled data. We utilize pre-trained self-supervised models as initialization for learning in a semi-supervised setting. The subplots provide the mean F-score of $10$ independent runs, where randomly selected instances are used to train the models. The bars with gray color represent the results of the networks trained only on the labeled instances while vertical black line shows results of fully-supervised model trained with entire data.}
\label{fig:ld}
\end{figure}

We next investigate the performance of our approach in a semi-supervised (or low-data) setting. For this purpose, we pre-train an encoder using unlabeled instances for each self-supervised task and utilize it as initialization for efficiently learning with few labeled instances on the end-task; for the end-task, we add a randomly-initialized dense layer with $1024$ hidden units before a linear output layer. The non-linear classifier is then learned and the encoder is fine-tuned with the specified number of instances per class. Specifically, for the defined auxiliary tasks and datasets, we use $5$ and $10$ examples for each category. We want to highlight that in a on-device learning case, a few labeled instances can be pooled from multiple users quite easily (e.g. $2$-$3$ examples per user) as compared to accumulating several hundred for learning fully-supervised models. Likewise, personalization can also be achieved through precisely asking for a few labels for targeted classes. In Figure~\ref{fig:ld}, we provide an average weighted F-score of $10$ independent experiment runs, comparing training from scratch (FS) with the pre-training as an effective initialization for learning a robust classifier. We show that in contrast to the purely supervised approach, leveraging unlabeled data for learning network parameters improves the performance on the end-task. Specifically, our self-supervised models greatly improve the F-score in the low-data setting, in some cases achieving F-scores nearly as good as networks trained with the entire labeled data. Similarly, the self-supervised trained models perform better than the autoencoder, which shows that, despite the simplicity, our proposed auxiliary tasks force the network to learn highly-generalizable features. For each experiment run, we randomly sample the stated number of annotated instances and use these to train all the networks, including fully-supervised baselines. 

On activity recognition, our methodology significantly improves the performance in low-data; for example, on the HHAR dataset with $5$ and $10$ instances, temporal shift and transformations tasks gain $4$ and $7$ points over the fully-supervised models' F-score of $0.60$ and $0.68$. respectively. Similarly, for MobiAct, pre-training with the temporal shift task helps achieve an F-score of $0.75$ ($5$ instances) and $0.82$ ($10$ instances), compared to $0.61$ and $0.73$ respectively for networks learned from scratch. Furthermore, we achieve identical improvements on UCI HAR, HAPT, and MotionSense with $5$ instances per class. The attained F-scores are $0.91$, $0.77$ and $0.83$ in contrast to $0.90$, $0.59$, and $0.77$ of fully-supervised models, respectively. Our method represents a $26$ points increase in F-score on the challenging problem of sleep stage scoring. Likewise, on physiological stress detection and device-free sensing problems, the benefit of pre-training with auxiliary tasks is further apparent, where the presented methods achieve $12$ points improvement in F-score over the baseline. These results suggest that self-supervision can greatly help with learning general-purpose representations that work well in the low-data regime. We also want to highlight that although the selection of an equal number of instances results in a balanced training set, we use the full test sets (as in earlier experiments) for evaluation, which could be imbalanced. Importantly, utilizing even bigger unlabeled datasets and combining weak-supervision methods can boost the quality of the learned representations. 

We emphasize that the broader objective of self-supervised methods is to learn high-level semantic features that can be used to solve an array of downstream tasks with minimal labeled data. The evaluation of our presented auxiliary tasks clearly highlights the benefit of pre-training the network with unlabeled data to achieve better generalization on the tasks of interest, with very few labeled instances. To the best of our knowledge, we, for the first time, evaluate self-supervised methods in a semi-supervised setting for problems involving multisensor data as earlier work developed fully-supervised network architectures or used classical autoencoding-based approaches for pre-training, followed by network fine-tuning with the entire labeled data. Overall, our approach provides a base for further work in developing sensing techniques that can achieve on-device personalization and perform continual, and few-shot learning, as the presented framework considerably reduces the requirement of labeled data from human annotators to learn the end-task models.

\subsubsection*{Effectiveness in a transfer learning setting}

\begin{figure}[t]
\subfloat[Autoencoder]{\includegraphics[width=5.5cm]{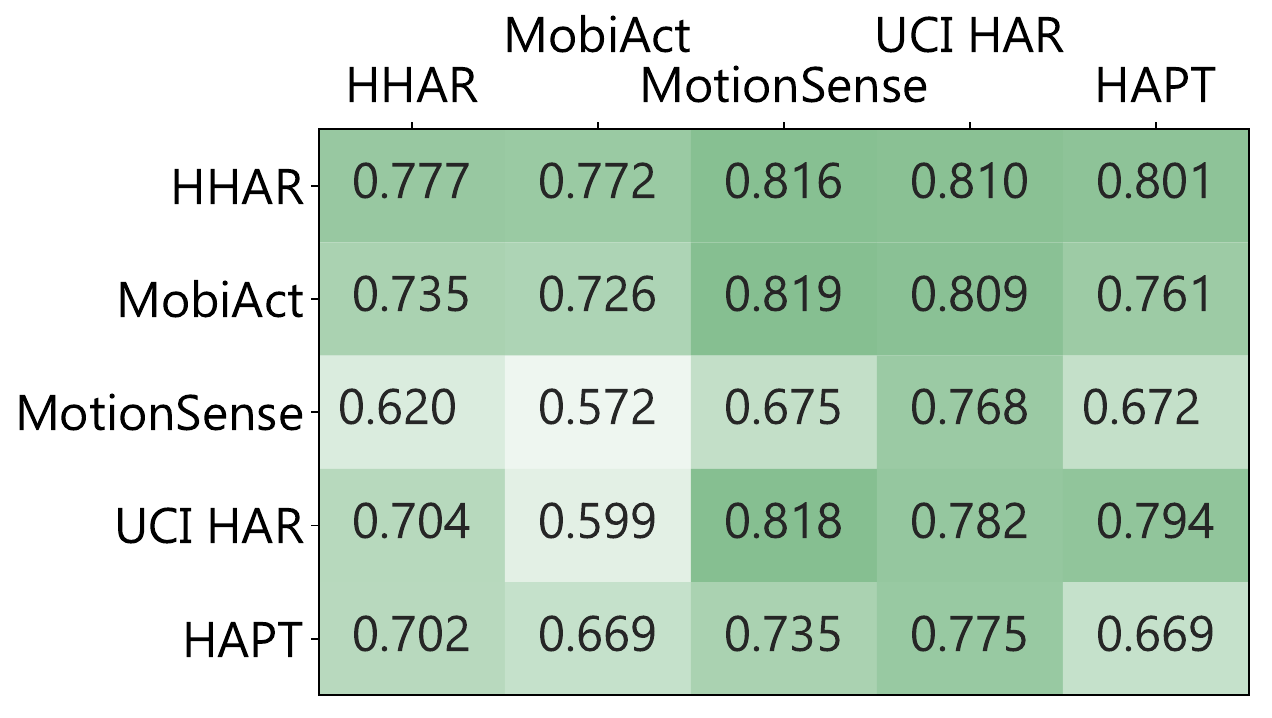}}  
\subfloat[Sensor Blend]{\includegraphics[width=4.2cm]{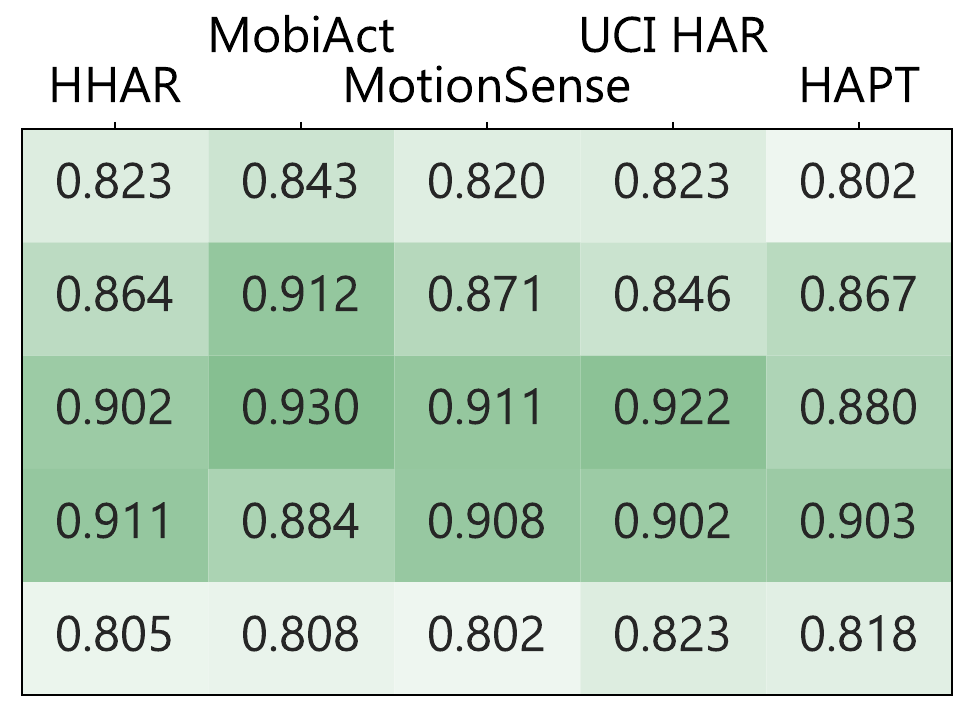}}
\subfloat[Fusion Magnitude]{\includegraphics[width=4.2cm]{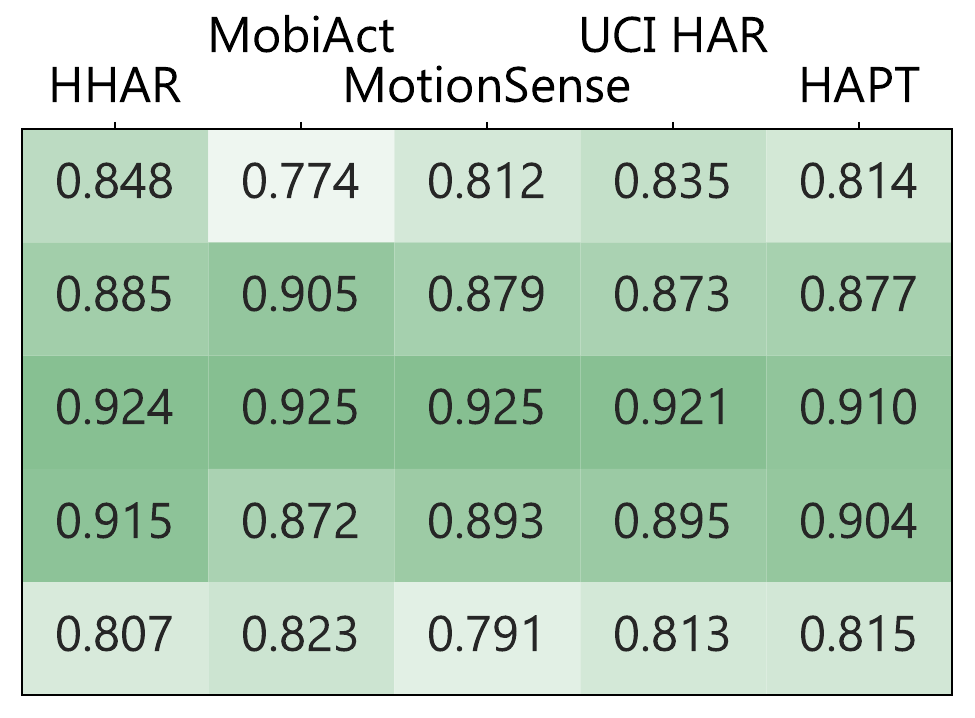}} \\
\subfloat[Feature Prediction]{\includegraphics[width=5.5cm]{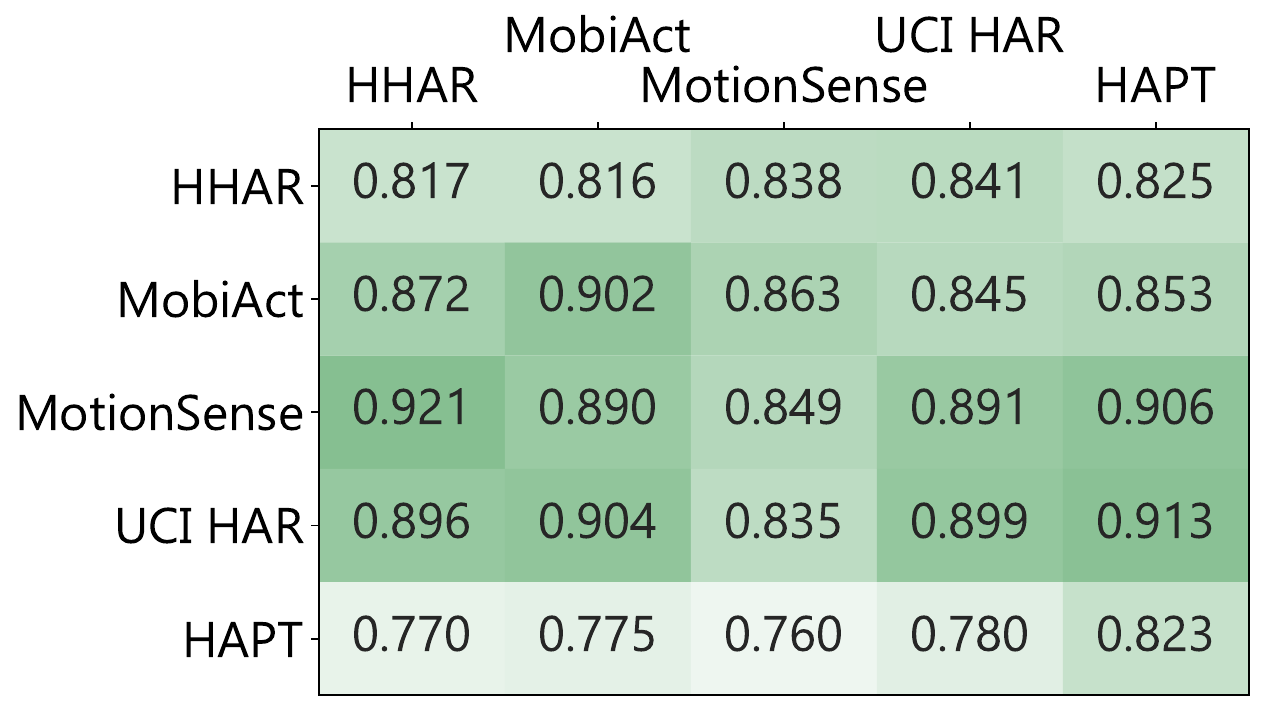}} 
\subfloat[Transformations]{\includegraphics[width=4.2cm]{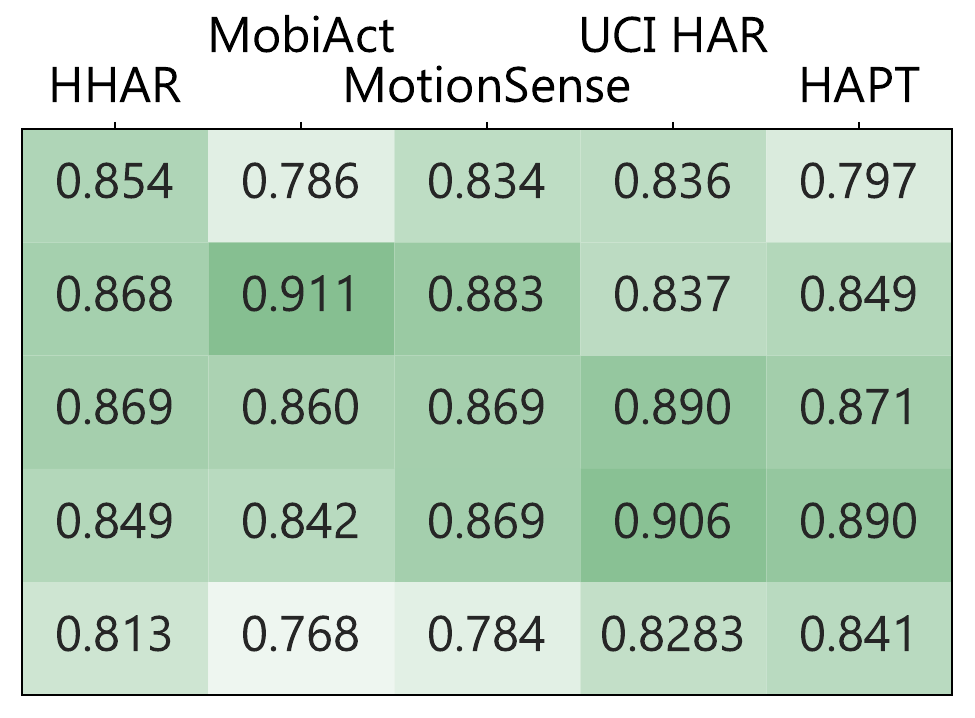}} 
\subfloat[Temporal Shift]{\includegraphics[width=4.2cm]{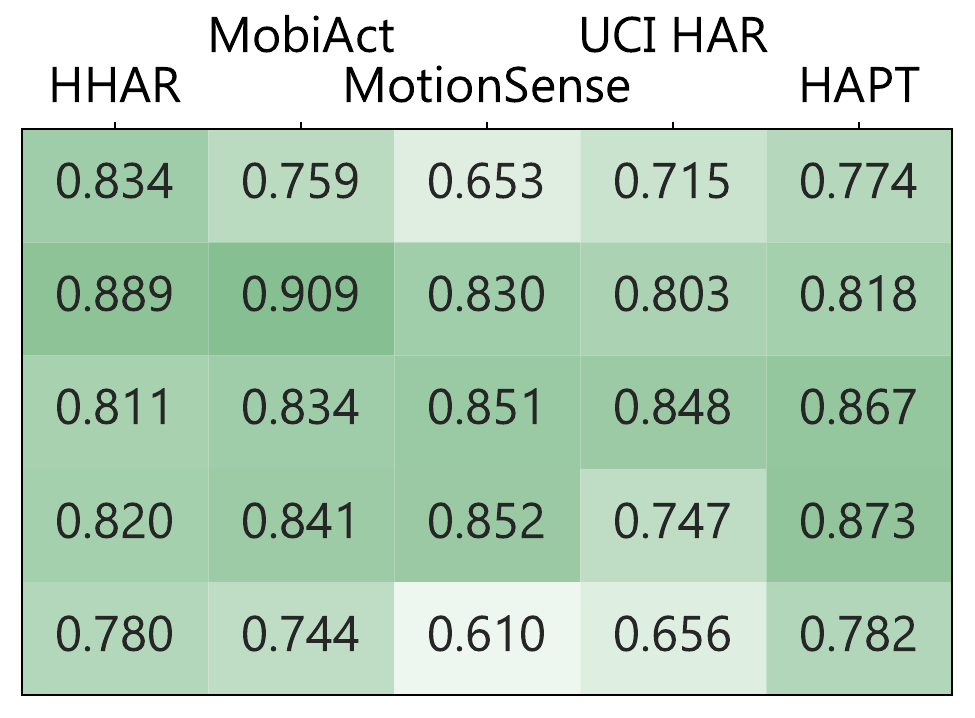}} \\
\subfloat[Modality Denoising]{\includegraphics[width=5.5cm]{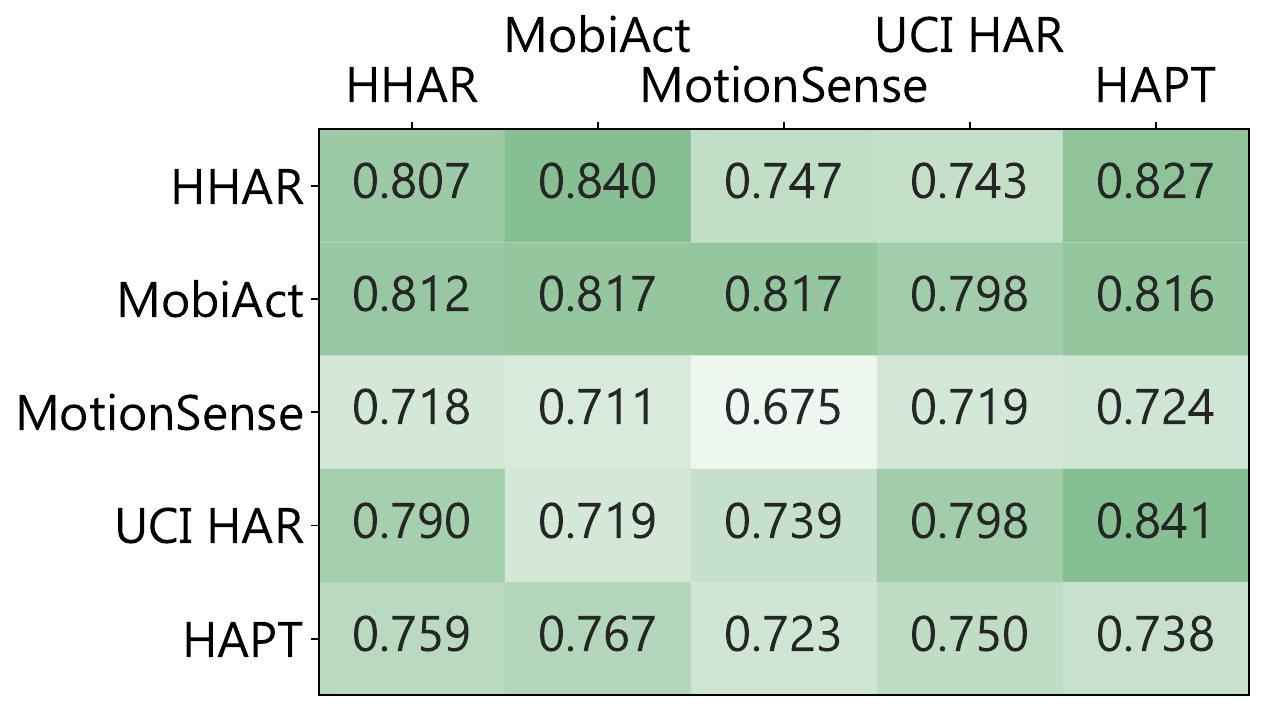}}
\subfloat[Odd Segment]{\includegraphics[width=4.2cm]{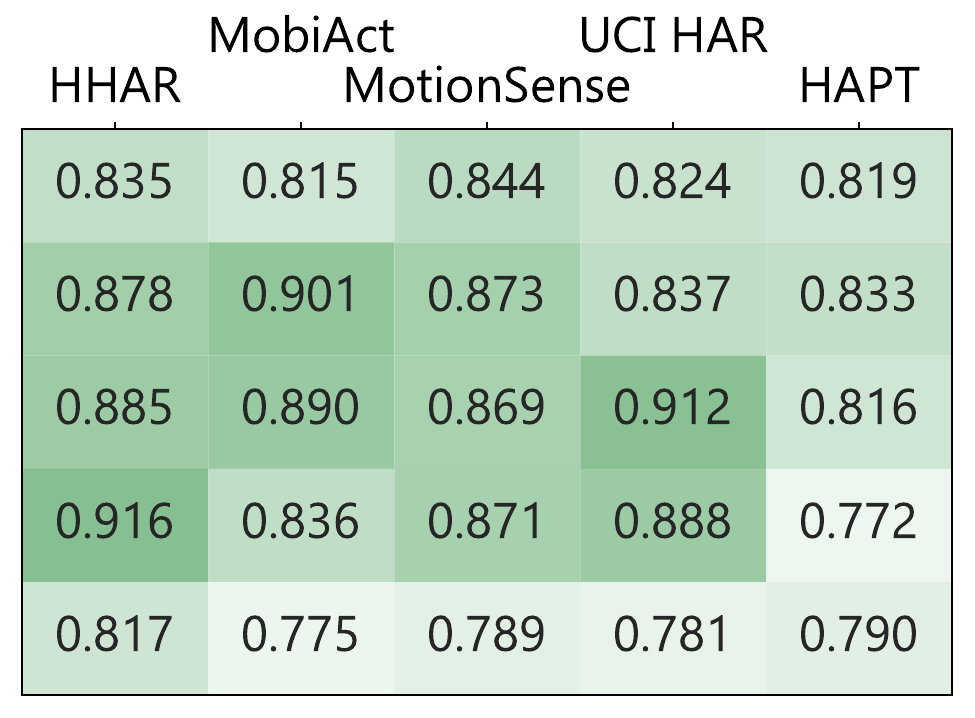}}
\subfloat[Triplet Loss]{\includegraphics[width=4.2cm]{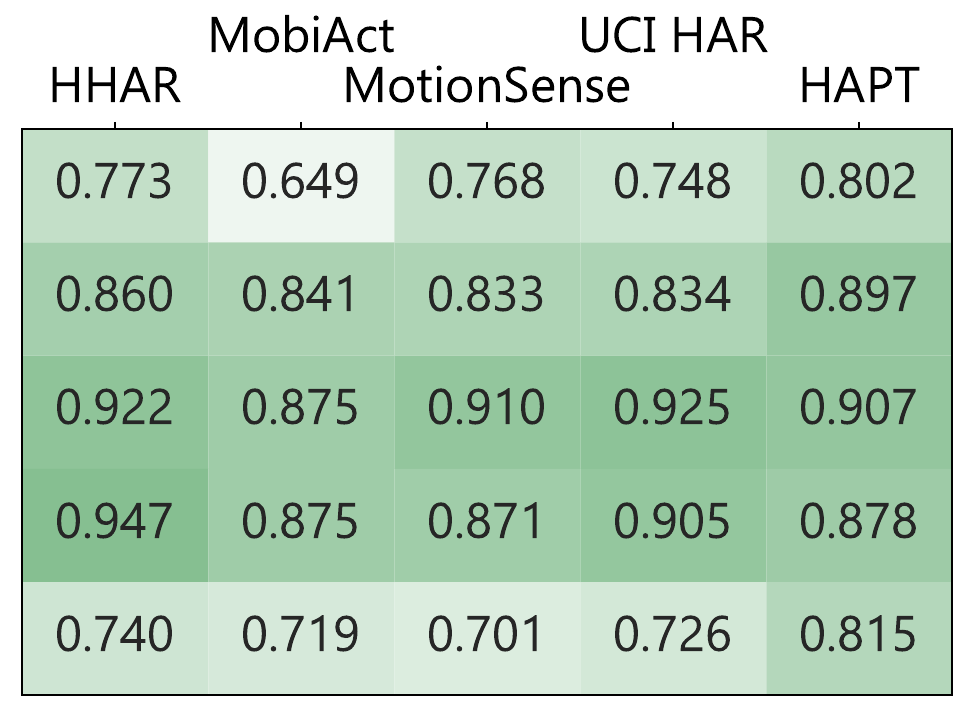}}
\caption{Generalization of the self-supervised representations under transfer learning setting. We evaluate the features transferability on activity recognition task by pre-training networks with each auxiliary task for every dataset. For solving downstream tasks, we train a linear classifier on-top of the frozen feature extractor $10$ times, independently, and report the average F-score. The diagonal entries denote the numbers when the source and target datasets are the same with the x-axis and y-axis representing target and source datasets, respectively.}
\label{fig:tf}
\end{figure}

In a real-world learning setup, there is a high chance that we are interested in a different dataset and downstream task than the one originating from the unlabeled data accessible for pre-training. A broadly useful auxiliary task is thus one that produces generalizable representations that transfer well to other related end tasks. To examine the transferability  property of the features learned with our proxy tasks, we evaluate their performance on the activity recognition datasets. To this end, we pre-train the feature extractor with each self-supervised objective (i.e. by discarding the semantic class labels) for all the five datasets (see section~\ref{sec:datasets}) and investigate their performance through a) training a linear classifier with the entire target annotated data and b) fine-tuning it end-to-end with few labeled data (i.e. learning an activity classifier with $5$ and $10$ instances of each class from target dataset). Figure~\ref{fig:tf} provides the results of the source-to-target transfer of self-supervised models trained with nine different auxiliary losses. The diagonal entries of each subplot represent the F-scores when the source and target datasets are the same. In comparison with autoencoder pre-training, features learned with our tasks transfer well between datasets. We observe that even leveraging smaller unlabeled datasets produces useful features, as with sensor-blend-task-learned features on UCI HAR scored $0.91$ F-score on the HHAR dataset. On the HAPT dataset of low input resolution (i.e. a segment size of $200$ samples) and complex postural activities, transfer learning improves the performance with approximately $8$ percentage points in F-score over pre-training on the same dataset. Importantly, our results are also competitive with the fully-supervised baselines on the respective datasets.    

\begin{figure}[htbp]
\subfloat[HHAR]{\includegraphics[width=8cm]{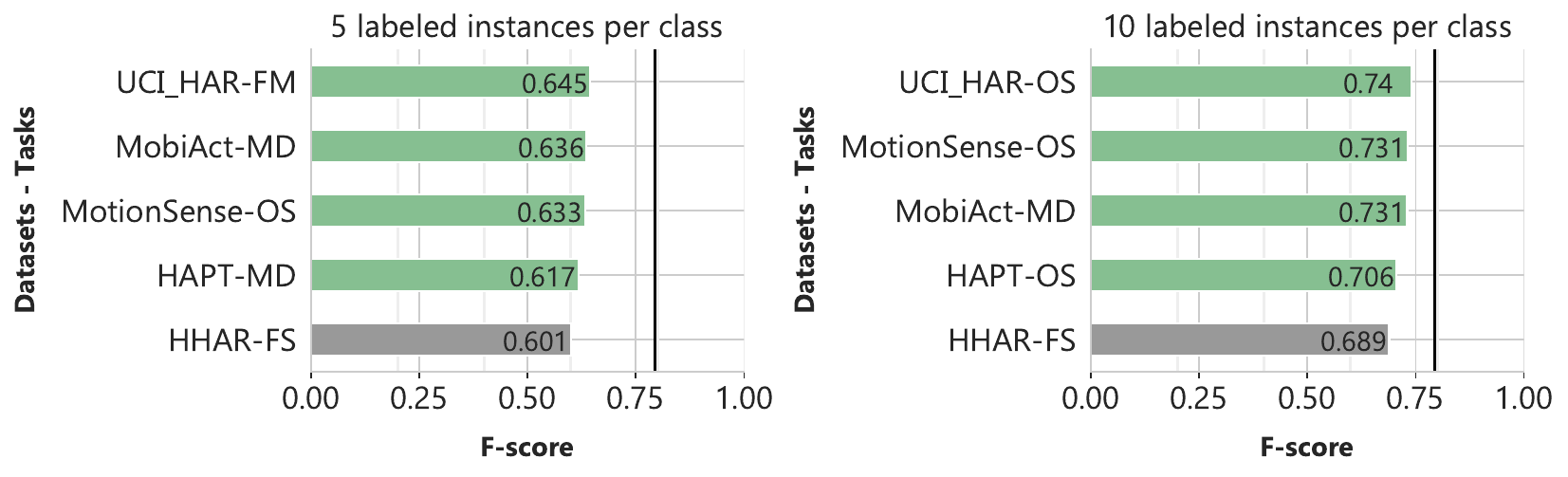}}  \\
\subfloat[MobiAct]{\includegraphics[width=8cm]{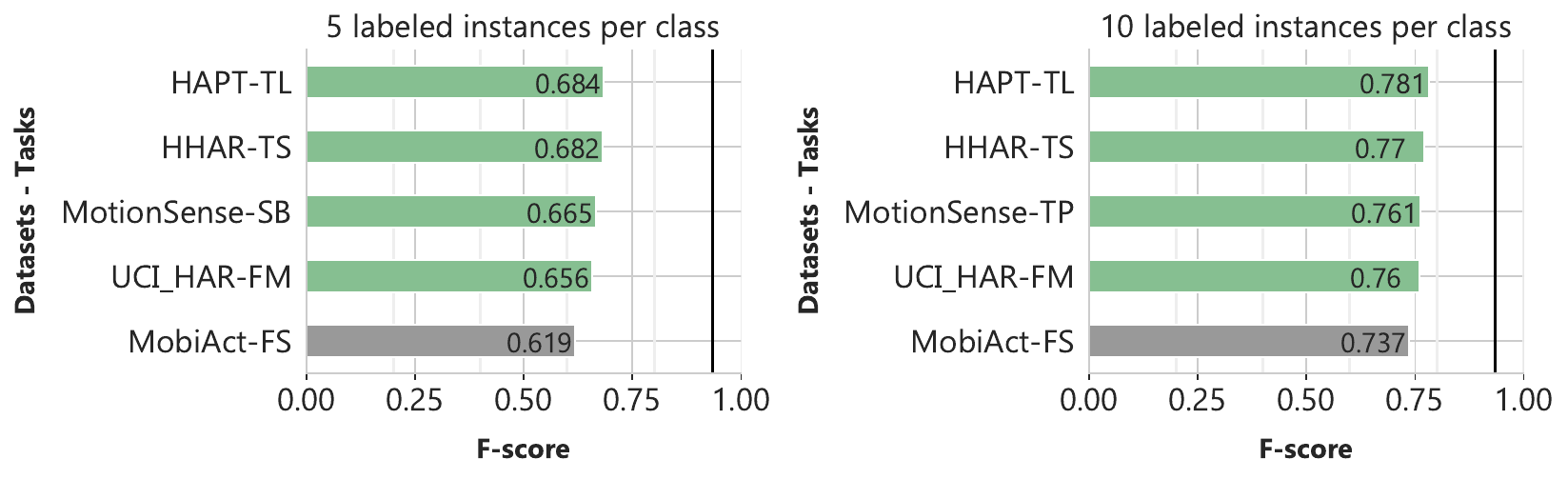}} \\
\subfloat[MotionSense]{\includegraphics[width=8cm]{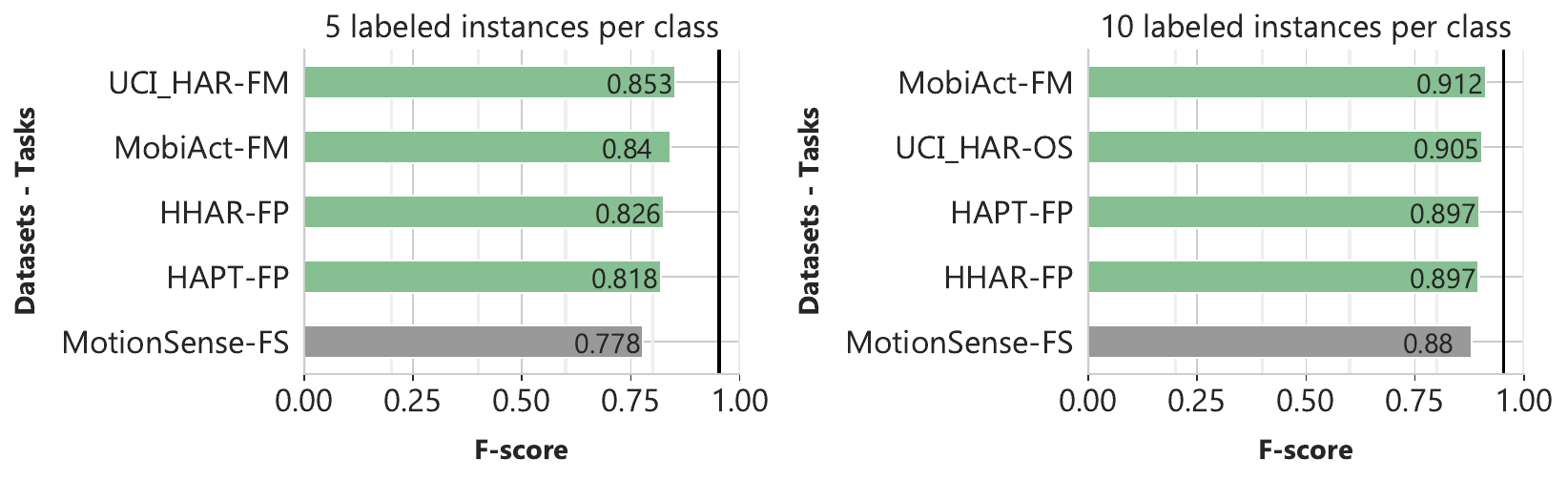}} \\
\subfloat[UCI HAR]{\includegraphics[width=8cm]{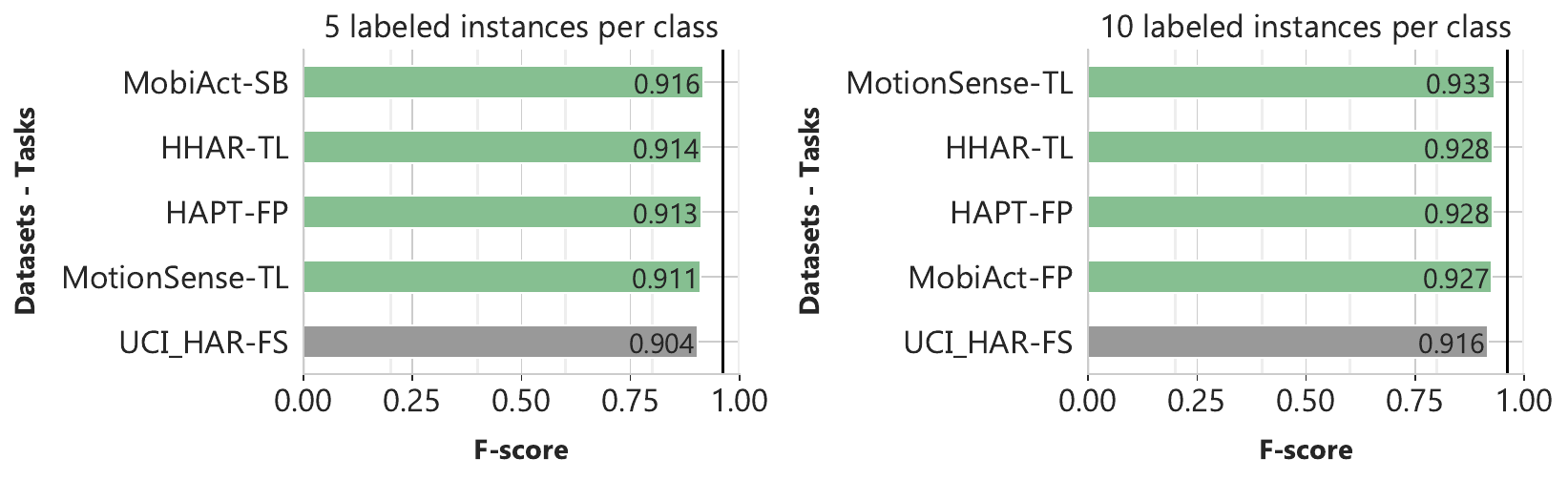}}  \\
\subfloat[HAPT]{\includegraphics[width=8cm]{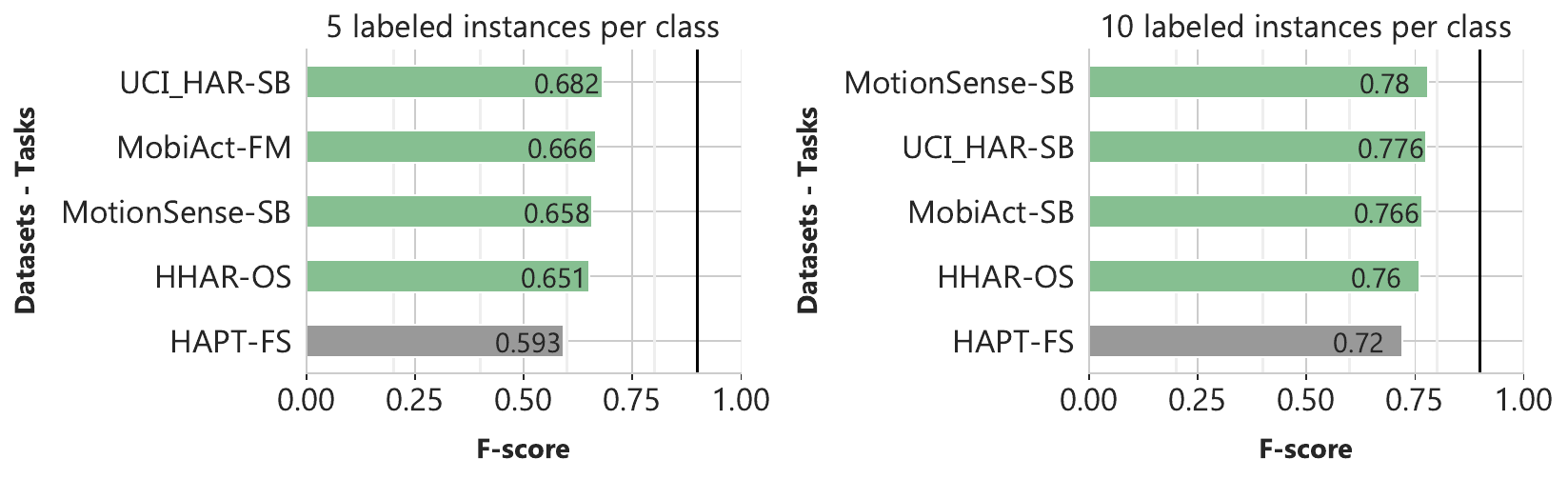}} \\
\subfloat{\includegraphics[width=6.5cm]{Figures/Low_Data/ld_caption.pdf}}
\caption{Contribution of self-supervised learning, and fine-tuning of the transferred networks in learning from few-data. We utilize a pre-trained model on each source data and train a non-linear classifier on the target task to assess the effectiveness of self-supervision for improving the recognition rate. The networks are fine-tuned with a specified number of instances per class $10$ times. For each source data, we provide mean results only of the best performing auxiliary task in order to improve readability.}
\label{fig:tf_ld}
\end{figure}

We further examine if the transferred self-supervised models are beneficial in learning from low-data; i.e. few labeled instances are available from the target data, but separate unannotated data is available for pre-training. We utilize the same network configuration as discussed earlier for low-data experiments and we fine-tune the model end-to-end. We randomly sample a specified number of instances and perform experiments $10$ times while utilizing the same instances for both types of networks (i.e. pre-trained and baseline) and report average F-score. In Figure~\ref{fig:tf_ld}, we present the results of optimal auxiliary tasks for each combination of the source to target transfer, where gray-colored bars show a fully-supervised baseline. Our experiments show that the features learned from different but related datasets do transfer well and improve the recognition rate even when as little as $5$ examples per class are available. On the MobiAct dataset, our approach with HAPT as source data results in an F-score of $0.68$ and $0.78$ compared to the training from scratch F-score of $0.61$ and $0.73$, respectively. Similarly, with HAPT as a target, transferring from the UCI HAR using the sensor blend task, the F-score improved from $0.59$ to $0.68$ and $0.72$ to $0.78$. Interestingly, on UCI HAR and MotionSense, the performance attained with our approach is very close to the purely supervised models trained with entirely labeled data (see Table~\ref{tab:linear}). 

Learning generalizable representations that can be reused for solving related tasks is an important property to have in a learning system. Our investigation of transferring unsupervised pre-trained models consistently highlights substantial performance improvements, indicating that the self-supervised features are broadly useful across different subjects, devices, environments and data collection protocols. In particular, the data efficiency enabled by our method in a low-data regime provides further evidence of semantic feature learning without merely over-fitting on the source dataset. It is also important to note that compared to earlier work which focuses on supervised transfer or joint-training on source and target datasets, we provide evaluation of unsupervised transfer and its ability to boost performance even with few-labeled data. Likewise, self-supervised learning has other benefits as it has been shown to improve adversarial robustness and uncertainty of deep models as compared to purely supervised methods~\cite{hendrycks2019using}. Although we did not study these aspects explicitly in this work, the results of transfer learning across domains hint that our auxiliary tasks also enhance the model's robustness; we leave an in-depth study for future work.

\subsubsection*{Cross-validation to determine robustness against subject variations}
\label{subsec:cv}
To validate the stability of our methodology against variations in subjects' data utilized for pre-training and downstream task evaluation, we perform $5$-fold cross-validation based on user split (i.e. the train and test division ($80-20$) is based on users with no overlap among them; train/test users are entirely independent); and we follow the same experimental setup as earlier. For each fold's data and surrogate task, we pre-train the models and train a linear classifier on top of the frozen network. The fully-supervised baseline is trained in an end-to-end manner, directly with the semantic labels. Table~\ref{tab:cv_linear} summarizes the results averaged across $5$ folds on eight considered datasets. We observe that the results achieved with self-supervision are consistent with earlier experiments. This highlights that our approach for sensory representation learning works well with different users' data and it is robust to subjects' differences. On the MobiAct dataset, the feature prediction and transformation recognition tasks achieve $0.90$ F-score, which is very close to a fully-supervised model's F-score of $0.91$. Likewise, on MIT DriverDb, self-supervision provides an impressive improvement over training from scratch. To summarize, these results suggest that the learned representations with unlabeled data learn useful features that can be used to a large extent for solving the end-task with a simple linear layer. Furthermore, we explore fine-tuning the last convolutional layer of the encoder while training a linear layer on downstream tasks. In Table~\ref{tab:cv_ft}, we show that fine-tuning a shared layer leads to a better performance than the fully-supervised model training from scratch on most of the datasets. The feature prediction task on the HHAR dataset achieved an F-score of $0.87$, which is $5$ points above the baseline. Likewise, on other datasets and tasks, our technique either bridges the gap or achieves broadly similar results as the supervised models. We think that careful fine-tuning of the architecture and related hyper-parameters could further improve the recognition rate of self-supervised networks. We note that a direct comparison of our approach with existing methods is not feasible as we learn representations from unlabeled data and evaluate through training a linear classifier, whereas, prior methods focus on fully-supervised learning with different architectures and evaluation strategies. However, to be comparative, we summarize related results here, which are only indicative. On MotionSense, our sensor blend task achieves an F-score of $0.92$ compared to $0.95$ and $0.86$ accuracy for trial- and subject-wise evaluation in~\cite{malekzadeh2018protecting}. For SleepEDF, our fusion magnitude task scores a kappa of $0.72$ compared to $0.76$ of a sophisticated fully-supervised model~\cite{supratak2017deepsleepnet}. Likewise, on WiFi sensing task, feature prediction proxy task results in an F-score of $0.85$ compared to the $0.90$ accuracy of an LSTM-based model~\cite{yousefi2017survey} over six classes.

We wondered whether pre-training with our auxiliary tasks is invariant to utilized subjects' data, as it is critical for learning in a real-world setting due to the non-curated nature of the data. We found that proxy tasks are highly stable and result in a similar performance as earlier, when a linear classifier is trained on top of self-supervised feature extractors. This analysis further shows that the self-supervised features are not necessarily subject-specific, but are general in nature.  Moreover, our evaluation demonstrates there is a room for improvement through selecting problem- or task-specific network architectures and using larger unlabeled datasets for unsupervised learning. Specifically, it would be valuable to explore unifying supervised and self-supervised objectives in a multi-task setting to personalize or adapt sensing models directly on user devices.

\begin{table}[htbp]
   \caption{Comparison of self-supervised representation learning to fully-supervised approach with $5$-fold cross-validation based on user-split. We pre-train the feature extractors for each fold's data and learn a linear classifier for the end-task as usual. We report weighted F-score averaged over the 5 folds, highlighting the robustness of our method to subject variations. See Table~\ref{tab:kappa_cv_linear} in appendix~\ref{appendix:kappa_results} for kappa scores.}\label{tab:cv_linear}
   \centering
   \subfloat{
    \small
    \centering
    \begin{tabular}{ccccc}
    Method & \textbf{HHAR} & \textbf{MobiAct} & \textbf{MotionSense} & \textbf{UCI HAR} \\ \hline
    Fully Supervised & 0.844$\pm$0.090 & 0.917$\pm$0.017 & 0.960$\pm$0.007 & 0.951$\pm$0.025 \\
    Random Init. & 0.199$\pm$0.047 & 0.394$\pm$0.086 & 0.284$\pm$0.086 & 0.268$\pm$0.208 \\
    Autoencoder & 0.722$\pm$0.085 & 0.736$\pm$0.021 & 0.752$\pm$0.050 & 0.831$\pm$0.041 \\ \hline
    Sensor Blend & 0.829$\pm$0.061 & 0.886$\pm$0.010 & \cellcolor[gray]{0.93}0.920$\pm$0.019 & \cellcolor[gray]{0.93}0.915$\pm$0.038 \\
    Fusion Magnitude & \cellcolor[gray]{0.93}0.841$\pm$0.040 & 0.889$\pm$0.014 & \cellcolor[gray]{0.93}0.924$\pm$0.025 & 0.899$\pm$0.049 \\
    Feature Prediction & 0.820$\pm$0.068 & \cellcolor[gray]{0.93}0.900$\pm$0.016 & 0.900$\pm$0.025 & 0.896$\pm$0.043 \\
    Transformations & \cellcolor[gray]{0.93}0.822$\pm$0.059 & \cellcolor[gray]{0.93}0.900$\pm$0.011 & 0.898$\pm$0.013 & \cellcolor[gray]{0.93}0.916$\pm$0.018 \\
    Temporal Shift & 0.811$\pm$0.057 & 0.890$\pm$0.017 & 0.889$\pm$0.027 & 0.793$\pm$0.030 \\
    Modality Denoise. & 0.798$\pm$0.077 & 0.834$\pm$0.029 & 0.780$\pm$0.058 & 0.829$\pm$0.056 \\
    Odd Segment & 0.812$\pm$0.079 & 0.890$\pm$0.015 & 0.901$\pm$0.014 & 0.861$\pm$0.015 \\
    Tripet Loss & 0.749$\pm$0.065 & 0.822$\pm$0.013 & 0.917$\pm$0.022 & 0.893$\pm$0.036 \\ \hline
    \end{tabular}
   } \hspace{0.02cm}
   \subfloat{
    \small
    \centering
    \begin{tabular}{ccccc}
    Method & \textbf{HAPT} & \textbf{Sleep-EDF} & \textbf{MIT DriverDb} & \textbf{WiFi CSI} \\ \hline
    Fully Supervised & 0.897$\pm$0.053 & 0.822$\pm$0.025 & 0.789$\pm$0.122 & 0.959$\pm$0.005 \\
    Random Init. & 0.155$\pm$0.061 & 0.072$\pm$0.021 & 0.206$\pm$0.015 & 0.214$\pm$0.044 \\
    Autoencoder & 0.818$\pm$0.064 & 0.701$\pm$0.026 & 0.850$\pm$0.054 & \cellcolor[gray]{0.93}0.793$\pm$0.014 \\ \hline
    Sensor Blend & 0.855$\pm$0.044 & 0.788$\pm$0.014 & 0.824$\pm$0.106 & - \\
    Fusion Magnitude & 0.840$\pm$0.040 & \cellcolor[gray]{0.93}0.795$\pm$0.025 & 0.859$\pm$0.061 & - \\
    Feature Prediction & \cellcolor[gray]{0.93}0.859$\pm$0.040 & 0.777$\pm$0.033 & 0.843$\pm$0.045 & \cellcolor[gray]{0.93}0.855$\pm$0.024 \\
    Transformations & \cellcolor[gray]{0.93}0.863$\pm$0.045 & 0.788$\pm$0.028 & 0.860$\pm$0.060 & 0.770$\pm$0.032 \\
    Temporal Shift & 0.837$\pm$0.042 & 0.753$\pm$0.027 & 0.844$\pm$0.082 & 0.729$\pm$0.015 \\
    Modality Denoise. & 0.835$\pm$0.050 & \cellcolor[gray]{0.93}0.797$\pm$0.029 & \cellcolor[gray]{0.93}0.864$\pm$0.061 & - \\
    Odd Segment & 0.821$\pm$0.043 & 0.767$\pm$0.037 & 0.839$\pm$0.071 & 0.793$\pm$0.018 \\
    Tripet Loss & 0.845$\pm$0.044 & 0.789$\pm$0.027 & \cellcolor[gray]{0.93}0.860$\pm$0.059 & 0.769$\pm$0.022 \\ \hline
    \end{tabular}
   }
\end{table}

\begin{table}[htbp]
   \caption{The effect of fine-tuning modality-agnostic encoder while learning downstream task under $5$-folds cross-validation as evaluated through weighted F-score. See Table~\ref{tab:kappa_cv_ft} in appendix~\ref{appendix:kappa_results} for kappa scores.}\label{tab:cv_ft}
   \centering
   \subfloat{
    \small
    \centering
    \begin{tabular}{ccccc}
    Method & \textbf{HHAR} & \textbf{MobiAct} & \textbf{MotionSense} & \textbf{UCI HAR} \\ \hline
    Fully Supervised & 0.844$\pm$0.090 & 0.917$\pm$0.017 & 0.960$\pm$0.007 & 0.951$\pm$0.025 \\
    Random Init. & 0.199$\pm$0.047 & 0.394$\pm$0.086 & 0.284$\pm$0.086 & 0.268$\pm$0.208 \\
    Autoencoder & 0.891$\pm$0.049 & 0.914$\pm$0.019 & 0.961$\pm$0.010 & 0.936$\pm$0.051 \\ \hline
    Sensor Blend & 0.893$\pm$0.062 & 0.919$\pm$0.011 & 0.964$\pm$0.011 & \cellcolor[gray]{0.93}0.949$\pm$0.036 \\
    Fusion Magnitude & 0.885$\pm$0.054 & 0.918$\pm$0.011 & 0.961$\pm$0.013 & 0.942$\pm$0.039 \\
    Feature Prediction & \cellcolor[gray]{0.93}0.894$\pm$0.050 & \cellcolor[gray]{0.93}0.930$\pm$0.014 & 0.962$\pm$0.003 & 0.943$\pm$0.047 \\
    Transformations & 0.893$\pm$0.052 & \cellcolor[gray]{0.93}0.933$\pm$0.0126 & \cellcolor[gray]{0.93}0.968$\pm$0.007 & 0.949$\pm$0.033 \\
    Temporal Shift & 0.885$\pm$0.055 & 0.920$\pm$0.014 & 0.941$\pm$0.012 & 0.915$\pm$0.050 \\
    Modality Denoise. & 0.886$\pm$0.061 & 0.929$\pm$0.015 & \cellcolor[gray]{0.93}0.966$\pm$0.011 & 0.933$\pm$0.054 \\
    Odd Segment & \cellcolor[gray]{0.93}0.894$\pm$0.067 & 0.927$\pm$0.011 & 0.962$\pm$0.004 & \cellcolor[gray]{0.93}0.951$\pm$0.030 \\
    Tripet Loss & 0.856$\pm$0.055 & 0.904$\pm$0.020 & 0.957$\pm$0.006 & 0.944$\pm$0.044 \\ \hline
    \end{tabular}
   } \hspace{0.02cm}
   \subfloat{
    \small
    \centering
    \begin{tabular}{ccccc}
    Method & \textbf{HAPT} & \textbf{Sleep-EDF} & \textbf{MIT DriverDb} & \textbf{WiFi CSI} \\ \hline
    Fully Supervised & 0.897$\pm$0.053 & 0.822$\pm$0.025 & 0.789$\pm$0.122 & 0.959$\pm$0.005 \\
    Random Init. & 0.155$\pm$0.061 & 0.072$\pm$0.021 & 0.206$\pm$0.015 & 0.214$\pm$0.044 \\
    Autoencoder & 0.883$\pm$0.059 & 0.764$\pm$0.028 & 0.804$\pm$0.132 & \cellcolor[gray]{0.93}0.911$\pm$0.032 \\ \hline
    Sensor Blend & \cellcolor[gray]{0.93}0.892$\pm$0.052 & 0.801$\pm$0.020 & 0.793$\pm$0.149 & - \\
    Fusion Magnitude & 0.884$\pm$0.051 & \cellcolor[gray]{0.93}0.808$\pm$0.023 & 0.788$\pm$0.148 & - \\
    Feature Prediction & 0.893$\pm$0.055 & 0.794$\pm$0.031 & 0.795$\pm$0.143 & 0.857$\pm$0.040 \\
    Transformations & \cellcolor[gray]{0.93}0.896$\pm$0.051 & \cellcolor[gray]{0.93}0.801$\pm$0.029 & 0.806$\pm$0.127 & 0.805$\pm$0.051 \\
    Temporal Shift & 0.890$\pm$0.052 & 0.781$\pm$0.027 & 0.805$\pm$0.133 & 0.758$\pm$0.048 \\
    Modality Denoise. & 0.882$\pm$0.051 & 0.796$\pm$0.028 & \cellcolor[gray]{0.93}0.858$\pm$0.051 & - \\
    Odd Segment & 0.888$\pm$0.048 & 0.778$\pm$0.035 & \cellcolor[gray]{0.93}0.849$\pm$0.050 & \cellcolor[gray]{0.93}0.854$\pm$0.032 \\
    Tripet Loss & 0.888$\pm$0.056 & 0.792$\pm$0.031 & 0.806$\pm$0.128 & 0.765$\pm$0.022 \\ \hline
    \end{tabular}
   }
\end{table}

\section{Impact and Limitations}
\label{sec:impact}
Our \textit{Sense and Learn} framework shows that it is possible to use unlabeled data, in addition to smaller amounts of labeled data, when learning features for varied classification problems. We believe our method is useful in practice, where obtaining labeled data is difficult and costly. Since the same approach, with a fixed neural network structure, provides gains for quite different application areas, ranging from activity recognition to sleep stage scoring, we also believe the method is applicable in practice. While it is true that a practitioner cannot be certain which self-supervised task will work best for a new application, the range of experiments we present should provide a valuable starting point as to which tasks are most promising. Moreover, our fine-tuning experiments (Table~\ref{tab:ft}) show that e.g. the Transformations task provides significant gains across all datasets even when using all available supervised data. Finally, self-supervised tasks don't need any labels while learning the representations, which opens up the possibility of using our framework for on-device Federated Learning~\cite{bonawitz2019towards}, where the sensor data never leaves the users' device (e.g., smartphone).

Self-supervised learning provides a scalable, inexpensive, and data efficient way to learn high-level features with deep neural networks without requiring strong labels, which could be unclear, noisy or limited for many real-world problems. However, there are limitations of these approaches which are also applicable to our methodology. First, deep neural networks are prone to learning via shortcuts through exploiting low-level cues in the input e.g. object textures and other local artifacts in image classification~\cite{geirhos2020shortcut}. The unintended cue learning is not limited to supervised methods, but is a problem for self-supervised methods too, as networks can use shortcuts to solve proxy task without learning anything useful (e.g. chromatic aberration in vision models~\cite{nathan2018improvements}). For time-series or multisensor inputs discovering, a model relying on shortcuts is an unsolved problem and could be challenging to detect. Second, as getting access to large unlabeled and labeled sensory datasets is difficult, evaluating how auxiliary tasks will perform on non-curated data or learning in an open-world environment needs further exploration. Third and last, interpretability and understanding the decision mechanism of deep models is another open area of research to address issues of model uncertainty, bias and fairness. The features learned with deep network could be non-interpretable, but we think that unifying shallow models using hand-crafted features with deep networks consuming raw input through knowledge distillation~\cite{hinton2015distilling} might shed light on the importance of certain features.

\section{Conclusion and Future Work}
\label{sec:conclusion}
We proposed a self-supervised framework for multisensor representation learning from unlabeled data, produced by the omnipresent sensors. To realize the vision of unsupervised learning for sensing systems and IoT in general, we developed eight novel auxiliary tasks that acquire their supervision signal directly from the raw input, without any human involvement. The defined proxy objectives are utilized to learn general and effective deep models for a wide variety of problems. Through extensive evaluation on eight publicly available datasets from four application domains, we demonstrate that the self-supervised networks learn useful semantic representations that are competitive with fully-supervised models (i.e. trained end-to-end with labeled data). In summary, we demonstrated that the straight-forward and computationally-inexpensive surrogate tasks perform well on downstream tasks of interest by learning a linear classifier on top of frozen feature extractors. We further showed that fine-tuning a pre-trained modality-agnostic encoder further improved the detection rate of a network. As the key objective of leveraging unannotated data is to reduce the labeled data required for the end-tasks, we have also shown that our approach significantly improves the performance in the low-data regime. In particular, with as few as $5$ to $10$ labeled examples per class, the self-supervised initialized networks achieve an F-score between $0.70$-$0.80$. Furthermore, we examined the effectiveness of learned representations in an unsupervised transfer setting with linear separability analysis and semi-supervised learning, achieving much better results than training from scratch.     

While in this work, we individually evaluate the quality of learned features for each auxiliary task, an interesting direction for future research is to jointly solve these problems in a multi-task setting, in order to learn more discriminative features. Likewise, an important area of investigation is to utilize the proposed tasks in a large-scale federated learning setting on distributed data. We believe this will truly highlight the potential of self-supervision for continual on-device (e.g., smartphones) learning and improving personalization. Finally, the general nature of our methodology offers the opportunity for leveraging self-supervision in other application areas, where labeled data accumulation is naturally difficult, such as arrhythmia detection.

\section*{Acknowledgements}
The authors would like to thank F\'elix de Chaumont Quitry, Marco Tagliasacchi and Richard F. Lyon for their valuable feedback and help with this work.
Various icons used in the figure are created by Sriramteja SRT, Berkah Icon, Ben Davis, Eucalyp, ibrandify, Clockwise, Aenne Brielmann, Anuar Zhumaev, and Tim Madle from the Noun Project.

\bibliographystyle{ACM-Reference-Format}
\bibliography{main}

\newpage
\appendix
\section{Appendix}
\subsection{Class Distribution}
\label{appendix:class_distribution}
\begin{figure}[!hbpt]
\subfloat[HHAR]{\includegraphics[width=3.9cm]{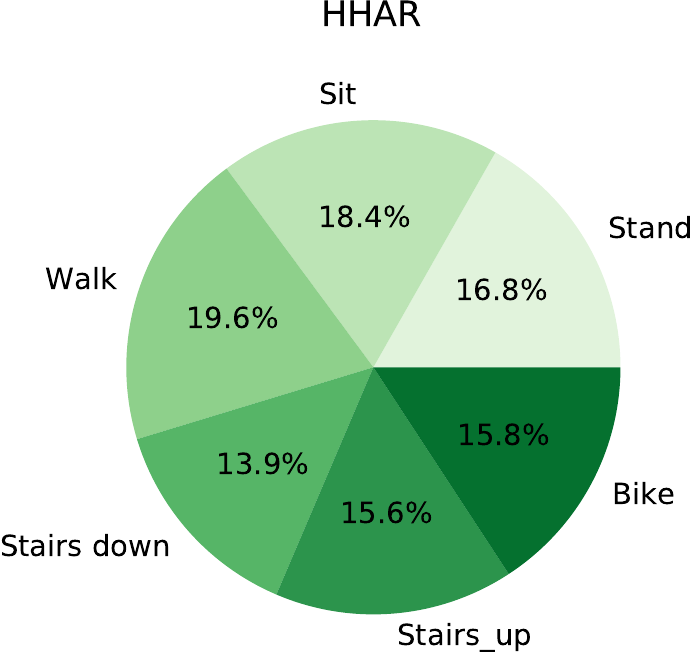}}\hspace{0.3cm}
\subfloat[MobiAct]{\includegraphics[width=3.5cm]{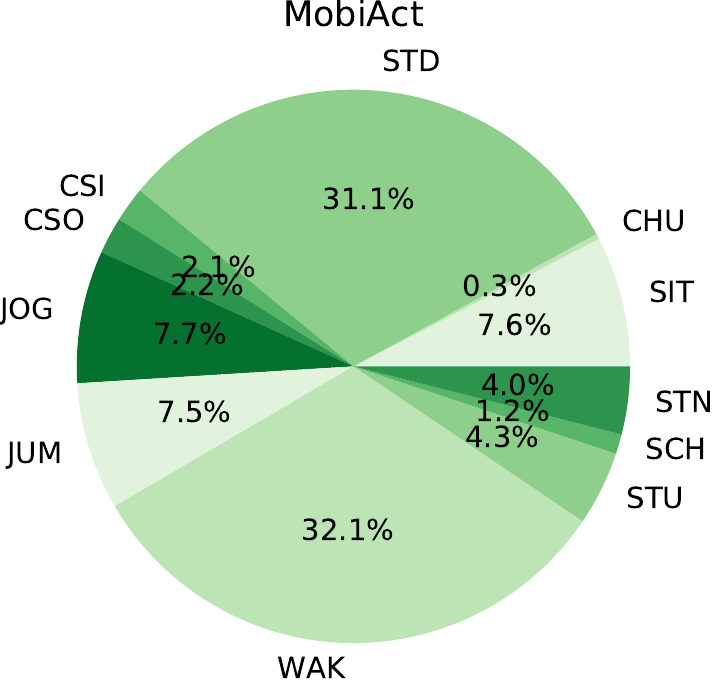}}\hspace{0.3cm}
\subfloat[MotionSense]{\includegraphics[width=4.2cm]{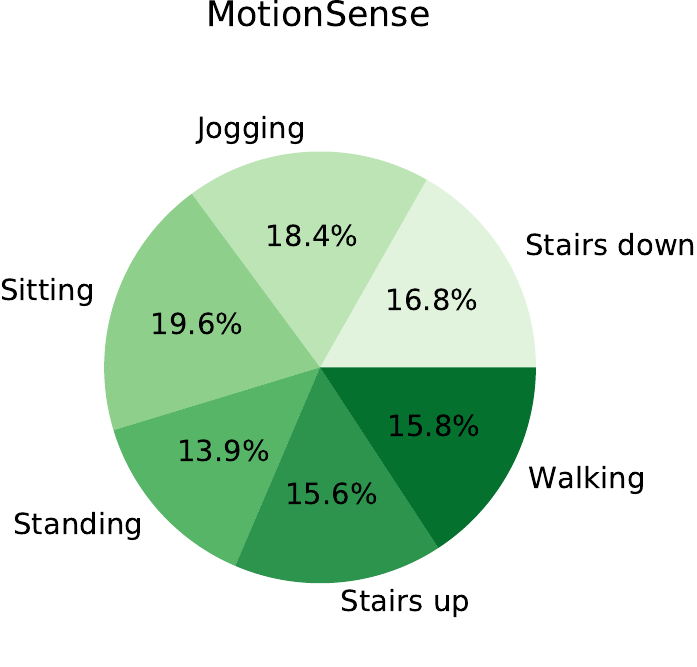}} \\
\subfloat[UCI HAR]{\includegraphics[width=4.2cm]{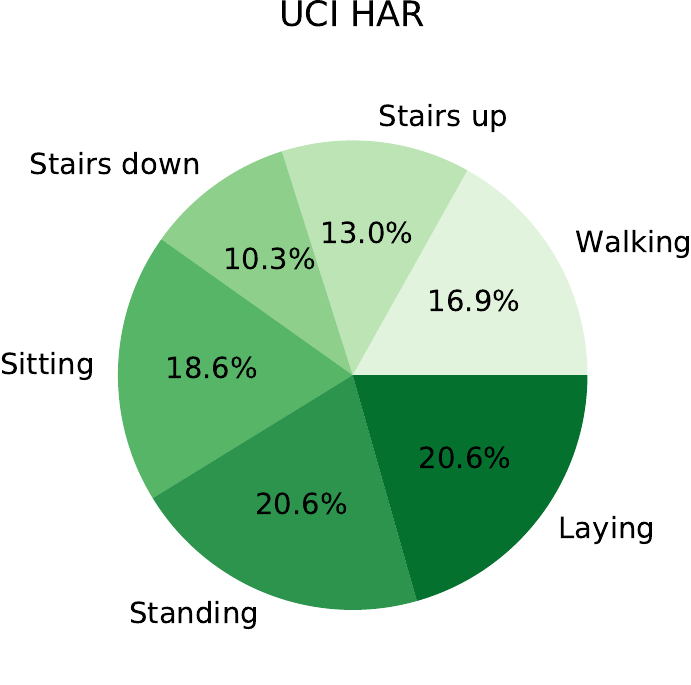}}\hspace{0.3cm}
\subfloat[Sleep-EDF]{\includegraphics[width=3.5cm]{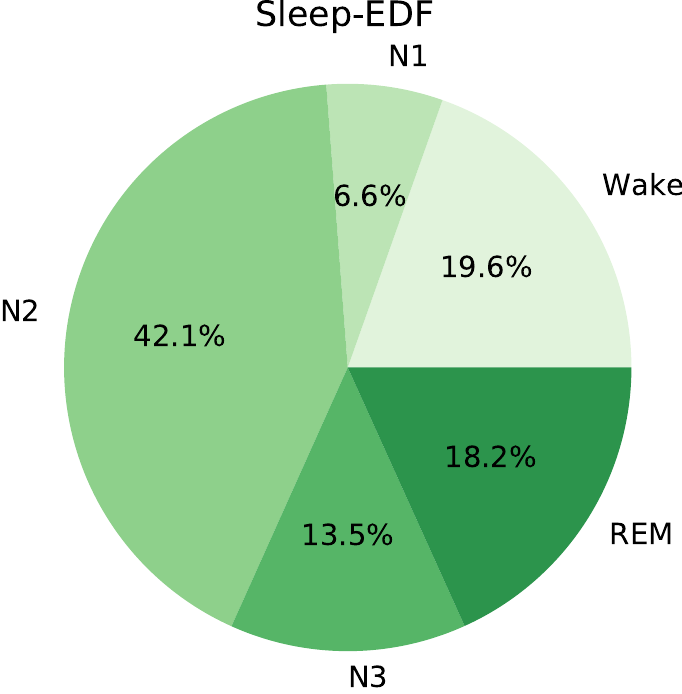}}\hspace{0.3cm} 
\subfloat[MIT DriverDb]{\includegraphics[width=3.5cm]{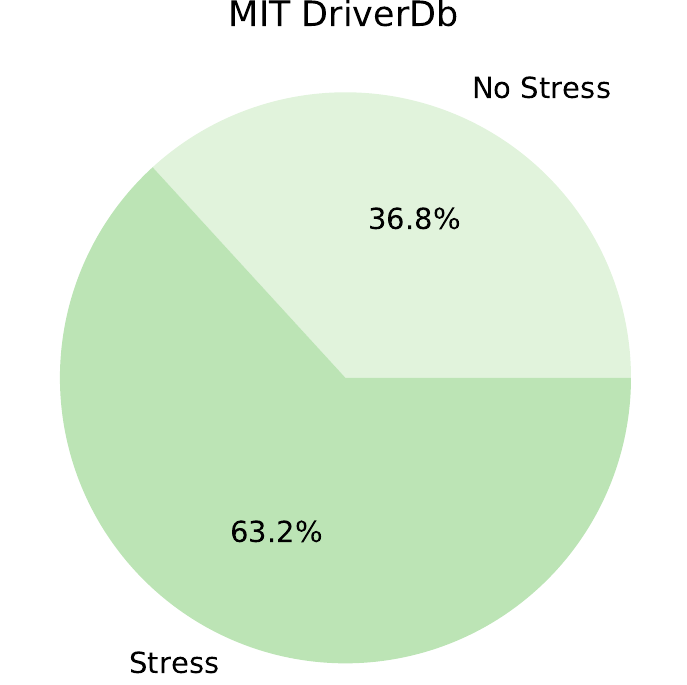}} \\
\subfloat[WiFi CSI]{\includegraphics[width=4cm]{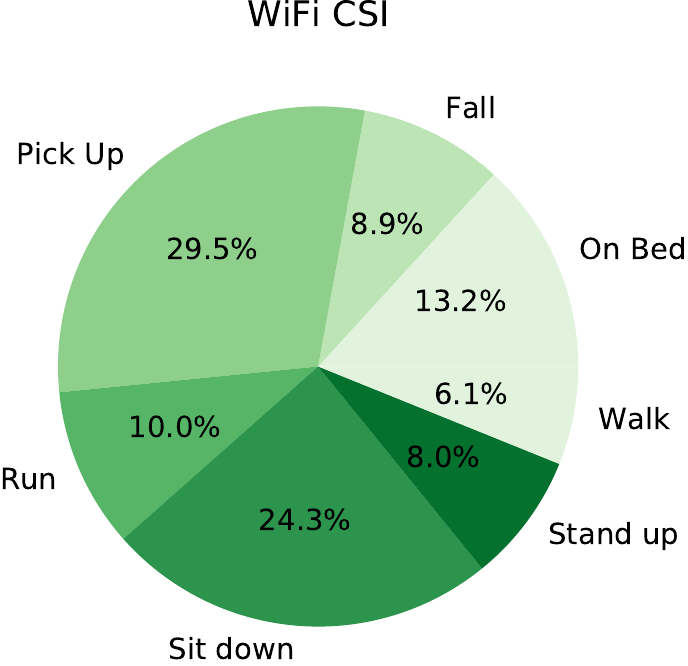}}\hspace{0.3cm} 
\subfloat[HAPT]{\includegraphics[width=4.8cm]{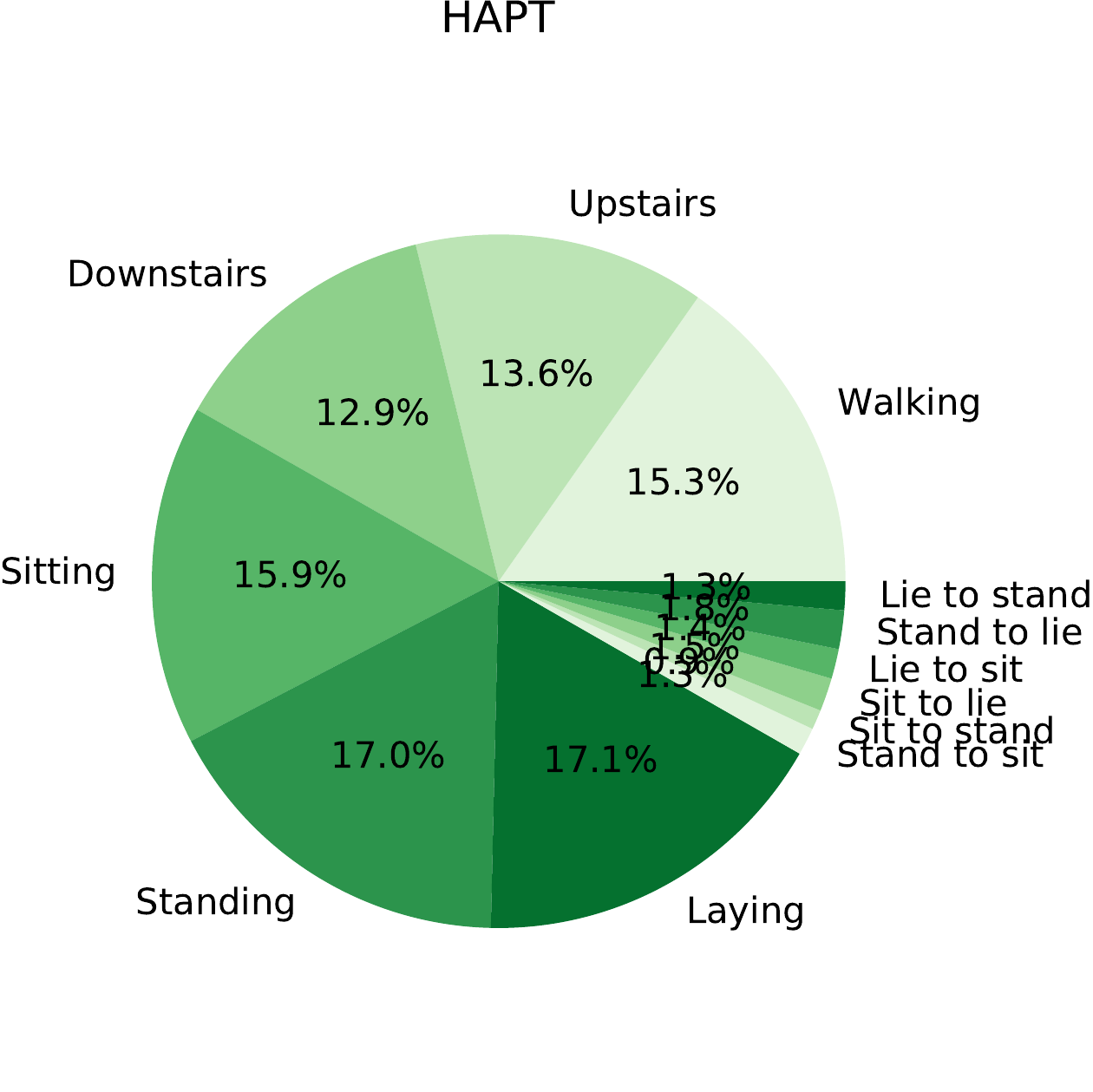}} 
\caption{Class distribution of the datasets used in evaluation.}
\label{fig:cd}
\end{figure}

\newpage

\subsection{Supplementary results with Cohen's kappa score}
\label{appendix:kappa_results}

\begin{table}[!htbp]
  \caption{Performance evaluation of self-supervised representations with a linear classifier. See Section~\ref{subsec:linear} for more details.}\label{tab:kappa_linear}
  \centering
  \subfloat{
    \scriptsize
    \centering
    \begin{tabular}{ccccc}
    Method & \textbf{HHAR} & \textbf{MobiAct} & \textbf{MotionSense} & \textbf{UCI HAR} \\ \hline
    Fully Supervised & 0.758$\pm$0.019 & 0.915$\pm$0.007 & 0.941$\pm$0.010 & 0.955$\pm$0.007 \\ 
    Random Init. & 0.115$\pm$0.069 & 0.254$\pm$0.122 & 0.153$\pm$0.086 & 0.157$\pm$0.104 \\
    Autoencoder & 0.732$\pm$0.004 & 0.696$\pm$0.002 & 0.654$\pm$0.011 & 0.749$\pm$0.041 \\ \hline
    Sensor Blend & 0.785$\pm$0.007 & \cellcolor[gray]{0.93}0.890$\pm$0.001 & \cellcolor[gray]{0.93}0.890$\pm$0.011 & 0.881$\pm$0.013 \\
    Fusion Magnitude & \cellcolor[gray]{0.93}0.815$\pm$0.006 & 0.880$\pm$0.002 & \cellcolor[gray]{0.93}0.907$\pm$0.014 & 0.874$\pm$0.013 \\
    Feature Prediction & 0.780$\pm$0.007 & 0.878$\pm$0.002 & 0.824$\pm$0.012 & 0.878$\pm$0.012 \\
    Transformations & \cellcolor[gray]{0.93}0.826$\pm$0.006 & \cellcolor[gray]{0.93}0.890$\pm$0.002 & 0.838$\pm$0.016 & \cellcolor[gray]{0.93}0.888$\pm$0.013 \\
    Temporal Shift & 0.801$\pm$0.010 & 0.884$\pm$0.004 & 0.818$\pm$0.019 & 0.708$\pm$0.027 \\
    Modality Denoise. & 0.771$\pm$0.007 & 0.789$\pm$0.004 & 0.656$\pm$0.017 & 0.758$\pm$0.043 \\
    Odd Segment & 0.801$\pm$0.008 & 0.877$\pm$0.002 & 0.837$\pm$0.015 & 0.871$\pm$0.010 \\
    Tripet Loss & 0.727$\pm$0.006 & 0.802$\pm$0.002 & 0.888$\pm$0.011 & \cellcolor[gray]{0.93}0.888$\pm$0.012 \\ \hline
    \end{tabular}
  } \hspace{0.01cm}
  \subfloat{
    \scriptsize
    \centering
    \begin{tabular}{ccccc}
    Method & \textbf{HAPT} & \textbf{Sleep-EDF} & \textbf{MIT DriverDb} & \textbf{WiFi CSI} \\ \hline
    Fully Supervised & 0.883$\pm$0.011 & 0.760$\pm$0.007 & 0.637$\pm$0.054 & 0.955$\pm$0.009 \\ 
    Random Init. & 0.041$\pm$0.039 & 0.026$\pm$0.068 & 0.077$\pm$0.206 & 0.012$\pm$0.042 \\
    Autoencoder & 0.646$\pm$0.004 & 0.566$\pm$0.014 & 0.736$\pm$0.005 & 0.713$\pm$0.005 \\ \hline
    Sensor Blend & 0.792$\pm$0.007 & 0.695$\pm$0.005 & 0.766$\pm$0.004 & - \\
    Fusion Magnitude & 0.789$\pm$0.005 & \cellcolor[gray]{0.93}0.700$\pm$0.008 & 0.771$\pm$0.010 & - \\
    Feature Prediction & \cellcolor[gray]{0.93}0.800$\pm$0.002 & 0.548$\pm$0.021 & 0.715$\pm$0.001 & \cellcolor[gray]{0.93}0.798$\pm$0.006 \\
    Transformations & \cellcolor[gray]{0.93}0.820$\pm$0.003 & 0.696$\pm$0.008 & \cellcolor[gray]{0.93}0.804$\pm$0.003 & 0.715$\pm$0.009 \\
    Temporal Shift & 0.753$\pm$0.004 & 0.599$\pm$0.014 & 0.751$\pm$0.011 & 0.670$\pm$0.013 \\
    Modality Denoise. & 0.717$\pm$0.003 & \cellcolor[gray]{0.93}0.702$\pm$0.002 & 0.792$\pm$0.003 & - \\
    Odd Segment & 0.758$\pm$0.004 & 0.689$\pm$0.004 & \cellcolor[gray]{0.93}0.758$\pm$0.004 & 0.722$\pm$0.009 \\
    Tripet Loss & 0.789$\pm$0.003 & 0.690$\pm$0.005 & 0.769$\pm$0.003 & 0.690$\pm$0.012 \\ \hline
    \end{tabular}
  } 
\end{table}

\begin{table}[!htbp]
  \caption{Improvement in recognition rate by fine-tuning the shared layers of the encoder while training on the end-task. See Section~\ref{subsec:linear} for more details.}\label{tab:kappa_ft}
  \centering
  \subfloat{
    \scriptsize
    \centering
    \begin{tabular}{ccccc}
    Method & \textbf{HHAR} & \textbf{MobiAct} & \textbf{MotionSense} & \textbf{UCI HAR} \\ \hline
    Fully Supervised & 0.758$\pm$0.019 & 0.915$\pm$0.007 & 0.941$\pm$0.010 & 0.953$\pm$0.01 \\
    Random Init. & 0.115$\pm$0.069 & 0.254$\pm$0.122 & 0.153$\pm$0.086 & 0.157$\pm$0.104 \\
    Autoencoder & 0.808$\pm$0.003 & 0.907$\pm$0.004 & 0.923$\pm$0.003 & 0.932$\pm$0.004 \\     \hline
    Sensor Blend & \cellcolor[gray]{0.93}0.815$\pm$0.011 & \cellcolor[gray]{0.93}0.927$\pm$0.005 & 0.921$\pm$0.006 & \cellcolor[gray]{0.93}0.948$\pm$0.004 \\
    Fusion Magnitude & 0.806$\pm$0.008 & 0.920$\pm$0.007 & 0.932$\pm$0.003 & 0.935$\pm$0.003 \\
    Feature Prediction & \cellcolor[gray]{0.93}0.816$\pm$0.008 & 0.919$\pm$0.003 & 0.940$\pm$0.004 & 0.931$\pm$0.003 \\
    Transformations & 0.802$\pm$0.007 & \cellcolor[gray]{0.93}0.932$\pm$0.005 & \cellcolor[gray]{0.93}0.940$\pm$0.006 & \cellcolor[gray]{0.93}0.944$\pm$0.007 \\
    Temporal Shift & 0.805$\pm$0.009 & 0.922$\pm$0.002 & 0.919$\pm$0.008 & 0.893$\pm$0.009 \\
    Modality Denoise. & 0.816$\pm$0.003 & 0.920$\pm$0.003 & 0.910$\pm$0.008 & 0.930$\pm$0.001 \\
    Odd Segment & 0.799$\pm$0.004 & 0.920$\pm$0.006 & 0.919$\pm$0.008 & 0.944$\pm$0.004 \\
    Tripet Loss & 0.806$\pm$0.014 & 0.886$\pm$0.008 & \cellcolor[gray]{0.93}0.944$\pm$0.004 & 0.940$\pm$0.003 \\ \hline
    \end{tabular}
  } \hspace{0.02cm}
  \subfloat{
    \scriptsize
    \centering
    \begin{tabular}{ccccc}
    Method & \textbf{HAPT} & \textbf{Sleep-EDF} & \textbf{MIT DriverDb} & \textbf{WiFi CSI} \\ \hline
    Fully Supervised & 0.883$\pm$0.011 & 0.760$\pm$0.007 & 0.637$\pm$0.054 & 0.955$\pm$0.009 \\
    Random Init. & 0.041$\pm$0.039 & 0.026$\pm$0.068 & 0.077$\pm$0.206 & 0.012$\pm$0.042 \\ 
    Autoencoder & 0.863$\pm$0.002 & 0.732$\pm$0.010 & 0.740$\pm$0.004 & \cellcolor[gray]{0.93}0.875$\pm$0.030 \\     \hline
    Sensor Blend & 0.880$\pm$0.003 & 0.739$\pm$0.005 & 0.748$\pm$0.029 & - \\
    Fusion Magnitude & 0.882$\pm$0.002 & 0.741$\pm$0.004 & 0.752$\pm$0.024 & - \\
    Feature Prediction & 0.878$\pm$0.003 & 0.652$\pm$0.010 & 0.702$\pm$0.006 & 0.791$\pm$0.048 \\
    Transformations & \cellcolor[gray]{0.93}0.882$\pm$0.003 & \cellcolor[gray]{0.93}0.755$\pm$0.005 & \cellcolor[gray]{0.93}0.767$\pm$0.011 & 0.783$\pm$0.0334 \\
    Temporal Shift & 0.857$\pm$0.008 & 0.696$\pm$0.004 & 0.751$\pm$0.011 & 0.678$\pm$0.070 \\
    Modality Denoise. & 0.868$\pm$0.004 & \cellcolor[gray]{0.93}0.752$\pm$0.007 & \cellcolor[gray]{0.93}0.764$\pm$0.003 & - \\
    Odd Segment & \cellcolor[gray]{0.93}0.883$\pm$0.003 & 0.730$\pm$0.007 & 0.691$\pm$0.048 & \cellcolor[gray]{0.93}0.828$\pm$0.037 \\
    Tripet Loss & 0.870$\pm$0.006 & 0.732$\pm$0.004 & 0.755$\pm$0.006 & 0.699$\pm$0.030 \\ \hline
    \end{tabular}
  } 
\end{table}

\begin{table}[!htbp]
  \caption{Comparison of self-supervised representation learning to fully-supervised approach with $5$-fold cross-validation based on user-split. See Section~\ref{subsec:cv} for more details.}\label{tab:kappa_cv_linear}
  \centering
  \subfloat{
    \scriptsize
    \centering
    \begin{tabular}{ccccc}
    Method & \textbf{HHAR} & \textbf{MobiAct} & \textbf{MotionSense} & \textbf{UCI HAR} \\ \hline
    Fully Supervised & 0.820$\pm$0.098 & 0.891$\pm$0.024 & 0.950$\pm$0.008 & 0.941$\pm$0.030 \\
    Random Init. & 0.107$\pm$0.072 & 0.272$\pm$0.084 & 0.202$\pm$0.082 & 0.190$\pm$0.223 \\
    Autoencoder & 0.672$\pm$0.104 & 0.703$\pm$0.029 & 0.719$\pm$0.058 & 0.805$\pm$0.041 \\  \hline
    Sensor Blend & 0.796$\pm$0.074 & 0.855$\pm$0.014 & \cellcolor[gray]{0.93}0.902$\pm$0.024 & \cellcolor[gray]{0.93}0.898$\pm$0.048 \\    
    Fusion Magnitude & \cellcolor[gray]{0.93}0.809$\pm$0.047 & 0.859$\pm$0.018 & \cellcolor[gray]{0.93}0.906$\pm$0.030 & 0.877$\pm$0.063 \\
    Feature Prediction & 0.787$\pm$0.083 & \cellcolor[gray]{0.93}0.876$\pm$0.020 & 0.878$\pm$0.030 & 0.875$\pm$0.051 \\
    Transformations & \cellcolor[gray]{0.93}0.789$\pm$0.071 & \cellcolor[gray]{0.93}0.876$\pm$0.015 & 0.873$\pm$0.017 & \cellcolor[gray]{0.93}0.900$\pm$0.022 \\
    Temporal Shift & 0.776$\pm$0.069 & 0.859$\pm$0.022 & 0.863$\pm$0.032 & 0.756$\pm$0.038 \\
    Modality Denoise. & 0.762$\pm$0.092 & 0.802$\pm$0.036 & 0.750$\pm$0.065 & 0.799$\pm$0.059 \\
    Odd Segment & 0.777$\pm$0.090 & 0.862$\pm$0.019 & 0.877$\pm$0.017 & 0.843$\pm$0.012 \\
    Tripet Loss & 0.707$\pm$0.077 & 0.777$\pm$0.018 & 0.897$\pm$0.027 & 0.873$\pm$0.043 \\ \hline
    \end{tabular}
  } \hspace{0.02cm}
  \subfloat{
    \scriptsize
    \centering
    \begin{tabular}{ccccc}
    Method & \textbf{HAPT} & \textbf{Sleep-EDF} & \textbf{MIT DriverDb} & \textbf{WiFi CSI} \\ \hline
    Fully Supervised & 0.880$\pm$0.063 & 0.760$\pm$0.037 & 0.577$\pm$0.219 & 0.949$\pm$0.006 \\
    Random Init. & 0.075$\pm$0.046 & 0.004$\pm$0.006 & 0.0$\pm$0.0 & 0.042$\pm$0.049 \\
    Autoencoder & 0.789$\pm$0.077 & 0.603$\pm$0.031 & 0.677$\pm$0.118 & \cellcolor[gray]{0.93}0.745$\pm$0.019 \\     \hline
    Sensor Blend & 0.832$\pm$0.053 & 0.715$\pm$0.026 & 0.636$\pm$0.203 & - \\
    Fusion Magnitude & 0.816$\pm$0.049 & \cellcolor[gray]{0.93}0.724$\pm$0.033 & 0.696$\pm$0.132 & - \\
    Feature Prediction & \cellcolor[gray]{0.93}0.837$\pm$0.048 & 0.701$\pm$0.034 & 0.661$\pm$0.100 & \cellcolor[gray]{0.93}0.821$\pm$0.029 \\
    Transformations & \cellcolor[gray]{0.93}0.842$\pm$0.051 & 0.712$\pm$0.026 & 0.698$\pm$0.134 & 0.716$\pm$0.039 \\
    Temporal Shift & 0.812$\pm$0.047 & 0.664$\pm$0.035 & 0.667$\pm$0.170 & 0.669$\pm$0.019 \\
    Modality Denoise. & 0.810$\pm$0.060 & \cellcolor[gray]{0.93}0.728$\pm$0.046 & \cellcolor[gray]{0.93}0.708$\pm$0.134 & - \\
    Odd Segment & 0.793$\pm$0.053 & 0.690$\pm$0.049 & 0.655$\pm$0.151 & 0.745$\pm$0.023 \\
    Tripet Loss & 0.820$\pm$0.052 & 0.715$\pm$0.037 & \cellcolor[gray]{0.93}0.698$\pm$0.128 & 0.715$\pm$0.027 \\ \hline
    \end{tabular}
  }
\end{table}

\begin{table}[!htbp]
  \caption{The effect of fine-tuning modality-agnostic encoder while learning downstream task under $5$-folds cross-validation. See Section~\ref{subsec:cv} for more details.}\label{tab:kappa_cv_ft}
  \centering
  \subfloat{
    \scriptsize
    \centering
    \begin{tabular}{ccccc}
    Method & \textbf{HHAR} & \textbf{MobiAct} & \textbf{MotionSense} & \textbf{UCI HAR} \\ \hline
    Fully Supervised & 0.820$\pm$0.098 & 0.891$\pm$0.024 & 0.950$\pm$0.008 & 0.941$\pm$0.030 \\
    Random Init. & 0.107$\pm$0.072 & 0.272$\pm$0.084 & 0.202$\pm$0.082 & 0.190$\pm$0.223 \\
    Autoencoder & 0.871$\pm$0.058 & 0.890$\pm$0.025 & 0.952$\pm$0.012 & 0.922$\pm$0.064 \\  \hline
    Sensor Blend & 0.873$\pm$0.072 & 0.895$\pm$0.015 & 0.956$\pm$0.014 & \cellcolor[gray]{0.93}0.939$\pm$0.043 \\     
    Fusion Magnitude & 0.864$\pm$0.063 & 0.893$\pm$0.016 & 0.951$\pm$0.016 & 0.930$\pm$0.047 \\
    Feature Prediction & \cellcolor[gray]{0.93}0.875$\pm$0.059 & \cellcolor[gray]{0.93}0.910$\pm$0.019 & 0.953$\pm$0.004 & 0.931$\pm$0.057 \\
    Transformations & 0.874$\pm$0.061 & \cellcolor[gray]{0.93}0.912$\pm$0.019 & \cellcolor[gray]{0.93}0.960$\pm$0.008 & 0.938$\pm$0.040 \\
    Temporal Shift & 0.864$\pm$0.065 & 0.895$\pm$0.020 & 0.926$\pm$0.017 & 0.900$\pm$0.059 \\
    Modality Denoise. & 0.865$\pm$0.072 & 0.909$\pm$0.020 & \cellcolor[gray]{0.93}0.957$\pm$0.013 & 0.919$\pm$0.066 \\
    Odd Segment & \cellcolor[gray]{0.93}0.875$\pm$0.078 & 0.906$\pm$0.016 & 0.953$\pm$0.005 & \cellcolor[gray]{0.93}0.940$\pm$0.037 \\
    Tripet Loss & 0.832$\pm$0.063 & 0.877$\pm$0.028 & 0.946$\pm$0.008 & 0.932$\pm$0.055 \\ \hline
    \end{tabular} 
  } \hspace{0.02cm}
  \subfloat{
    \scriptsize
    \centering
    \begin{tabular}{ccccc}
    Method & \textbf{HAPT} & \textbf{Sleep-EDF} & \textbf{MIT DriverDb} & \textbf{WiFi CSI} \\ \hline
    Fully Supervised & 0.880$\pm$0.063 & 0.760$\pm$0.037 & 0.577$\pm$0.219 & 0.949$\pm$0.006 \\
    Random Init. & 0.075$\pm$0.046 & 0.004$\pm$0.006 & 0.0$\pm$0.0 & 0.042$\pm$0.049 \\
    Autoencoder & 0.864$\pm$0.070 & 0.688$\pm$0.043 & 0.603$\pm$0.252 & \cellcolor[gray]{0.93}0.891$\pm$0.039 \\     \hline
    Sensor Blend & \cellcolor[gray]{0.93}0.875$\pm$0.061 & 0.730$\pm$0.031 & 0.591$\pm$0.268 & - \\
    Fusion Magnitude & 0.866$\pm$0.060 & \cellcolor[gray]{0.93}0.741$\pm$0.034 & 0.585$\pm$0.263 & - \\
    Feature Prediction & 0.876$\pm$0.065 & 0.726$\pm$0.033 & 0.596$\pm$0.257 & 0.824$\pm$0.048 \\
    Transformations & \cellcolor[gray]{0.93}0.880$\pm$0.059 & \cellcolor[gray]{0.93}0.732$\pm$0.041 & 0.610$\pm$0.239 & 0.764$\pm$0.054 \\
    Temporal Shift & 0.873$\pm$0.061 & 0.709$\pm$0.032 & 0.603$\pm$0.254 & 0.703$\pm$0.070 \\
    Modality Denoise. & 0.864$\pm$0.060 & 0.727$\pm$0.043 & \cellcolor[gray]{0.93}0.694$\pm$0.112 & - \\
    Odd Segment & 0.870$\pm$0.058 & 0.705$\pm$0.054 & \cellcolor[gray]{0.93}0.677$\pm$0.111 & \cellcolor[gray]{0.93}0.826$\pm$0.036 \\
    Tripet Loss & 0.869$\pm$0.067 & 0.724$\pm$0.043 & 0.608$\pm$0.242 & 0.709$\pm$0.031 \\ \hline
    \end{tabular}
  }
\end{table}

\end{document}